\documentclass[lettersize,journal]{IEEEtran}
\usepackage{amsmath,amsfonts}
\usepackage{algorithmic}
\usepackage{algorithm}
\usepackage{array}
\usepackage[caption=false,font=normalsize,labelfont=sf,textfont=sf]{subfig}
\usepackage{textcomp}
\usepackage{stfloats}
\usepackage{url}
\usepackage{verbatim}
\usepackage{graphicx}
\usepackage{cite}
\usepackage{threeparttable}

\usepackage{subfiles} %
\usepackage{makecell} %
\usepackage{multirow} %
\usepackage{subcaption}
\usepackage{color,xcolor}
\usepackage{booktabs}
\usepackage{makecell}
\usepackage{colortbl}
\usepackage{pifont}%
\usepackage{algorithm} %
\usepackage{algorithmic} %
\usepackage[switch]{lineno}
\usepackage{amssymb} %
\usepackage{fontawesome}
\usepackage[misc]{ifsym}

\definecolor{shapecolor}{rgb}{0.0,0.5,0.0}
\newcommand{\name}{DAVSSM}%

\definecolor{arylideyellow}{rgb}{0.91, 0.84, 0.42}

\definecolor{cvprblue}{rgb}{0.21,0.49,0.74}
\usepackage[pagebackref,breaklinks,colorlinks,citecolor=cvprblue]{hyperref}

\newcommand{\tabincell}[2]{\begin{tabular}{@{}#1@{}}#2\end{tabular}}

\begin{document}

\title{Preserving Full Degradation Details for Blind Image Super-Resolution}



\author{Hongda Liu, Longguang Wang, Ye Zhang, Kaiwen Xue, Shunbo Zhou, Yulan Guo}

\markboth{Journal of \LaTeX\ Class Files,~Vol.~14, No.~8, August~2021}%
{Shell \MakeLowercase{\textit{et al.}}: A Sample Article Using IEEEtran.cls for IEEE Journals}


\maketitle

\begin{abstract}
The performance of image super-resolution relies heavily on the accuracy of degradation information, especially under blind settings. Due to the absence of true degradation models in real-world scenarios, previous methods learn distinct representations by distinguishing different degradations in a batch. However, the most significant degradation differences may provide shortcuts for the learning of representations such that subtle difference may be discarded. In this paper, we propose an alternative to learn degradation representations through reproducing degraded low-resolution (LR) images. By guiding the degrader to reconstruct input LR images, full degradation information can be encoded into the representations. In addition, we develop a distribution alignment loss to facilitate the learning of the degradation representations by introducing a bounded constraint. {Moreover, to achieve larger receptive fields to capture information from a wider region for better restoration results, we introduce a degradation-aware Mamba module to efficiently model long-range dependency between the anchor pixel and the surrounding informative pixels. And the module strikes a flexible adaption to various degradations based on the learned representations.} Experiments show that our representations can extract accurate and highly robust degradation information. Evaluations on both synthetic and real images demonstrate that our ReDSR achieves state-of-the-art performance for the blind SR tasks.
\end{abstract}

\begin{IEEEkeywords}
Image Super-Resolution, \and Preserving Degradation Information, \and Distribution Alignment Loss, \and State Space Model.
\end{IEEEkeywords}

\section{Introduction}

Single image super-resolution (SR) aims at reconstructing a high-resolution (HR) image from its low-resolution (LR) counterpart. As a typical inverse problem, SR is highly coupled with the degradation model~\cite{ref32_internalgan}. In early stages, most CNN-based methods~\cite{ref33_SRCNN,ref34_RCAN,ref35_SRRRN,ref36_VDSR} are developed based on an assumption that the degradation is known and fixed (\textit{e.g.}, bicubic downsampling).

To ease the ill-posedness of the SR task under diverse degradations, numerous works first conduct degradation estimation and then use it as priori information for SR~\cite{ref32_internalgan,ref14_ikc,ref37_DAN,ref38_MANet}. However, these methods are sensitive to the estimated degradation and the estimation error may be magnified by the SR network to produce severe artifacts. Inspired by contrastive learning~\cite{ref39_contrastive1,ref40_contrastive2}, Wang \emph{et al.}~\cite{ref3_dasr} proposed to distinguish different degradations rather than explicitly estimate the degradation. Xia \emph{et al.}~\cite{ref5_kdsr} further improved the discriminability of degradation representations by employing knowledge distillation. However, the learned degradation representations exhibit strong bias towards significant difference and cannot well capture the subtle degradation difference, resulting in limited generalization performance.

 \begin{figure}[t]
		\centering
		\includegraphics[width=1.0\linewidth]{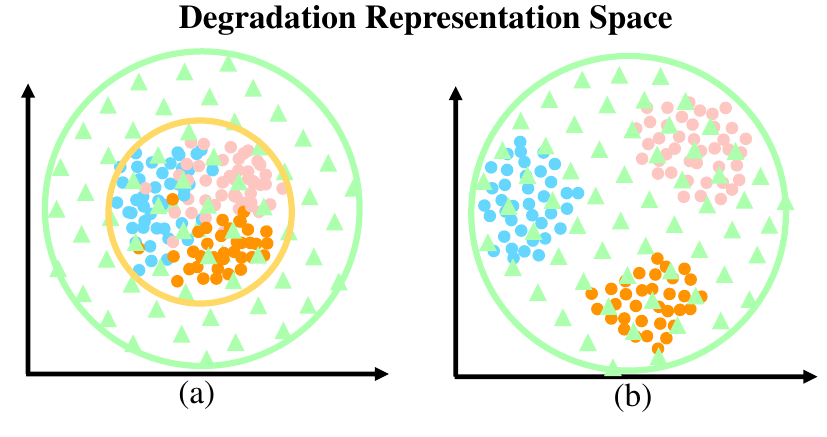}
		\caption{An illustration of degradation representation space. Note that, the triangles represent samples in a pre-defined distribution while the circles represent extracted degradation representations. (a)  Representations learned without a bounded constraint (\emph{e.g.}, DASR~\cite{ref3_dasr}). (b) Representations learned with a bounded constraint (Ours).}
		\label{distribution_fig}
\end{figure}

\begin{figure*}[t]
		\centering
		\includegraphics[width=1.0\linewidth]{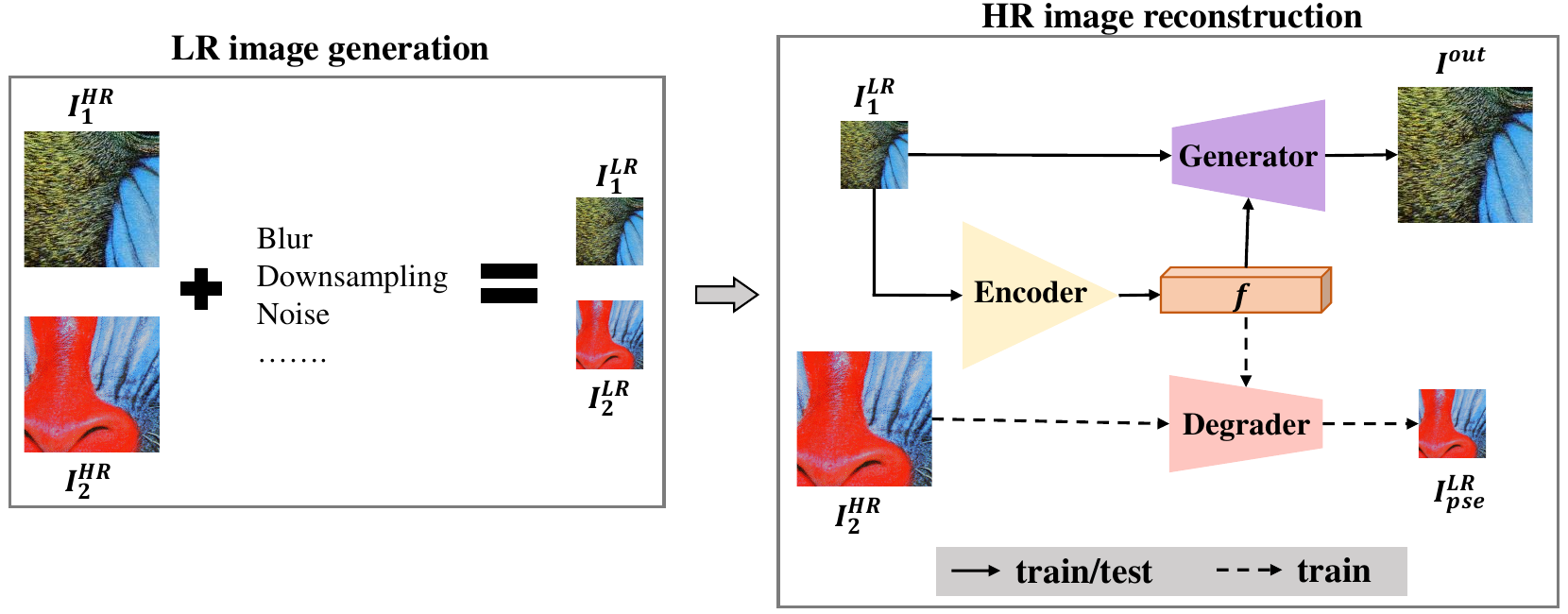}
		\caption{An overview of our ReDSR framework.}
		\label{Framework}
\end{figure*}

In this paper, we propose an alternative to learn degradation representations by reproducing degraded LR images. Specifically, the LR image is first fed to an {encoder} to extract degradation representation. Then, this degradation representation is leveraged by a {degrader} to reproduce the LR image from the HR image. By reproducing the LR image, full degradation information details can be captured in a compact degradation representation. With degradation information being encoded, degradation representation is passed to the generator to super-resolve the input LR image (Fig.~\ref{Framework}). {Existing re-degradation methods~\cite{ref41_KOALANET,ref42_REBLUR} only consider a single degradation type (\emph{e.g.}, blur) and cannot be extended to real-world applications where images are corrupted by multiple degradations. In contrast, our high-dimension degradation representation encodes multiple degradations in an implicit way. Moreover, we introduce a distribution alignment loss to bound the learned representations in a pre-defined distribution (\emph{e.g.}, Gaussian distribution).} As compared to previous approaches that learn representation in an unbounded space (Fig.~\ref{distribution_fig}(a)), our distribution alignment loss facilitates more distinctive representations to be extracted (Fig.~\ref{distribution_fig}(b)), especially for degradations unseen in the training set~\cite{ref43_CVAE}.


{In addition to distinctive degradation representations, it has been demonstrated that effective receptive field plays a crucial role in SR. Different degradations require various sizes of the receptive field to generate suitable SR results. For more complex degradations, larger receptive fields are often required to leverage more context information to facilitate the reconstruction of the anchor pixel~\cite{ref13_mambair}. }
Despite promising performance, previous methods~\cite{ref44_SWINIR,ref45_HAT} achieve larger receptive fields at the cost of higher computational cost. Recently, a novel State Space Model (SSM) called Mamba~\cite{ref11_MAMBA} is proposed in the NLP field for long sequence modeling with linear complexity~\cite{ref46_hungry,ref11_MAMBA,ref47_efficientlySSM,ref48_longSSM,ref49_simplified}. Mamba introduces an effective solution to balance large receptive field and computational efficiency~\cite{ref8_zhuvisionmamba,ref7_liuvmamba,ref13_mambair,ref50_umamba}. 
Inspired by Mamba, we propose a novel Degradation-aware Vision State Space Module (DAVSSM). Particularly, we propose a novel Degradation-aware Selective Scan Structured State Space Sequence Block (DS6 block), which efficiently introduces degradation information to state space updating by predicting weighting parameters in SSM from degradation representations. Moreover, we introduce a zigzag selective scanning method to process image token sequences in a spatially continuous way, which improves semantic continuity. In addition, to ease local pixel forgetting and channel redundancy introduced by Mamba, we develop a Degradation-condition Modulation Module (DCMM), which incorporates various degradation information to perform flexible feature adaption.

In summary, our contributions are four-fold:
\begin{itemize}
    \item We propose an alternative to extract degradation information from LR images by learning degradation representation to guide the degrader to reproduce the input LR images.
    \item We introduce a distribution alignment loss to facilitate the learning of discriminative representations by constraining the representations in a bounded pre-defined space.
    \item We design a pluggable degradation-aware Mamba module with flexible adaption to different degradations based on learned degradation representations. In addition, a zigzag scanning method is introduced to obtain superior SR results.
    \item Extensive experiments show that our method produces state-of-the-art performance on benchmark datasets.
\end{itemize}

\section{Related Work}

\subsection{Blind Image Super-Resolution}

Blind image super resolution aims to super-resolve LR images with unknown degradations. Early methods~\cite{ref32_internalgan,ref14_ikc,ref99_VDIR} commonly follow a two-step pipeline that first estimate the degradation model and then conduct image SR conditioned on the degradation. 
Specifically, Gu \emph{et al.}~\cite{ref14_ikc} proposed an iterative kernel correction (IKC) method to alternately correct estimated degradation and conduct image SR. Huang \emph{et al.}~\cite{ref37_DAN} developed a deep alternating network (DAN) by iteratively estimating the degradation and
restoring the SR image. Liang \emph{et al.}~\cite{ref38_MANet} proposed a mutual affine network (MANet) to exploit the interdependence between different channels by mutual affine transformation. Since numerous iterations are required to obtain accurate degradation information at test time, these methods are usually time-consuming. Recently, some works like BSRGAN~\cite{ref2_bsrgan} and Real-ESRGAN~\cite{ref1_realesrgan} design more complex and comprehensive degradation processes to better cover the real-world degradation space. The complex degradation combinations are hard to be explicitly estimated by a degradation estimator.

Inspired by the developments of contrastive learning~\cite{ref39_contrastive1,ref40_contrastive2}, several efforts~\cite{ref3_dasr,ref51_CDSR,ref52_DAASR} have been made to leverage contrastive learning to extract discriminative representations to implicitly obtain degradation information. Wang \emph{et al.}~\cite{ref3_dasr} first introduced degradation representation learning to distinguish different degradations in the representation space rather than explicit degradation estimation. Zhou \emph{et al.}~\cite{ref51_CDSR} proposed content-aware embedding to encode more information into the representations. Xia \emph{et al.}~\cite{ref5_kdsr} proposed to employ knowledge distillation to further improve the discriminability of degradation representations in a two-stage pipeline.

\subsection{Energy-Based Model}

Energy-based model (EBM) has demonstrated great advantages in modeling low-dimensional data distributions for image generation~\cite{ref56_Lecuntutorial,ref55_kumar,ref54_hadselldimension,ref57_Zhaoenergy,ref53_du2019implicit}. Early EBMs~\cite{ref68_AckleyBoltzmann,ref69_Hintoncontrastivedivergence,ref70_SalakhutdinovDeepBoltzmann,ref71_SalakhutdinovEfficient} formulated the energy function as a composition of latent and observable variables. Later EBMs~\cite{ref59_mnihlearning,ref58_Hintontraining,ref60_ranzatoefficient} directly mapped image samples to the representations in a equilibrium distribution. However, the number of samples limits the quality of generated images~\cite{ref60_ranzatoefficient} and the model is not robust to the variance created by sampling~\cite{ref58_Hintontraining}. To remedy this, Zhao \emph{et al.}~\cite{ref72_ZhaoEBGAN} combines GAN~\cite{ref73_GAN} and Auto-Encoder~\cite{ref74_AE} to achieve image quality improvement. {Previous works commonly minimize L1/L2 distance between the generated images and target images for data modeling. Nevertheless, these methods suffer slow convergence and low image quality~\cite{ref53_du2019implicit}.} Furthermore, visual data, for example, while represented in a high-dimensional space, often exists on a low-dimensional manifold~\cite{ref64_generativematching}. Inspired by this, variational auto encoder (VAE)~\cite{ref66_VAE,ref43_CVAE} is proposed to explicitly match low-dimensional data representation space to Gaussian distribution space by minimizing Kullback-Leibler (KL) divergence~\cite{ref67_KLdivergence} distance.

Recently, Gretton \emph{et al.}~\cite{ref61_grettonkernel} proposed an empirical estimate of Maximum Mean Discrepancy (MMD) to measure the distance between two distributions. 
Compared with KL divergence, MMD preforms more stably and consistently in low-dimensional data distribution modeling~\cite{ref65_Gritsenkospectral,ref64_generativematching}.
Later, Rizzo \emph{et al.}~\cite{ref62_rizzoenergydistance} further simplified the MMD to an energy distance by removing the kernel trick that is scalable to multi-dimensional representation space and easier to be implemented~\cite{ref63_Goldenberg}.

\subsection{State Space Model}

State Space Model (SSM), as a key component in control theory, is recently introduced to deep learning as a competitive backbone for state space transforming~\cite{ref47_efficientlySSM,ref49_simplified}. Compared to the quadratic complexity of the self-attention mechanism, SSM achieves competitive performance in long sequence modeling with only linear complexity. Structured State Space Sequence model (S4)~\cite{ref47_efficientlySSM} proposes to normalize the parameter matrices into a diagonal structure, which is a seminal work for the deep state space model in modeling the long-range dependency. Furthermore, S5 layer~\cite{ref49_simplified} is proposed based on S4 and introduces MIMO SSM and efficient parallel scan.~\cite{ref46_hungry} designs H3 layer which nearly fills the performance gap between SSM and Transformer attention in natural language modeling.~\cite{ref48_longSSM} builds the Gated State Space layer on S4 by introducing more gating units to boost the expressivity and accelerate model training. To tackle the limitation of S4 in capturing the contextual information, Gu \emph{et al.}~\cite{ref11_MAMBA} propose Mamba, which is a novel parameterization method for SSM that integrates an input-dependent selection scan mechanism (referred to as S6) and efficient hardware design. Mamba outperforms Transformer on natural language and enjoys linear scaling with input length. Moreover, there are also pioneering works that adopt SSM to vision tasks such as image classification~\cite{ref8_zhuvisionmamba,ref7_liuvmamba}, image restoration~\cite{ref13_mambair,ref75_Vmambair}, biomedical image segmentation~\cite{ref50_umamba,ref76_mambaunet2} and others~\cite{ref78_s4nd,ref77_sstransformer,ref79_ssvideo}.

\section{Methodology}

\subsection{Preliminaries}

Structured state space sequence models (S4) and Mamba are inspired by the continuous system, which maps a 1-D function or sequence $x(t)\in\mathbb{R}{\rightarrow}y(t)\in\mathbb{R}$ through an implicit latent state $h(t)\in\mathbb{R}^N$. Concretely, continuous-time SSMs can be formulated as linear ordinary differential equations (ODEs) as follows,
\begin{equation}\label{pure_ssm}
\begin{split}
  h'(t) &= \mathbf{A} h(t) + \mathbf{B} x(t),  \\
  y(t)  &= \mathbf{C} h(t) + \mathbf{D} x(t).
\end{split}
\end{equation}
where $\mathbf{A} \in \mathbb{R}^{N\times N}$, $\mathbf{B}\in \mathbb{R}^{N\times 1}$, $\mathbf{C}\in \mathbb{R}^{1\times N}$ and $\mathbf{D} \in \mathbb{R}$ are the weighting parameters.

After that, the discretization process is typically adopted to integrate Eq.~\ref{pure_ssm} into practical deep learning algorithms. The process  introduces a timescale parameter $\mathbf{\Delta}$ to transform the continuous parameters $\mathbf{A}$, $\mathbf{B}$ to discrete parameters $\mathbf{\overline{A}}$, $\mathbf{\overline{B}}$. The commonly used method for transformation is zero-order hold (ZOH), which is defined as follows:
\begin{equation}
\begin{aligned}
\label{eq:zoh}
\mathbf{\overline{A}} &= \exp{(\mathbf{\Delta}\mathbf{A})}, \\
\mathbf{\overline{B}} &= (\mathbf{\Delta} \mathbf{A})^{-1}(\exp{(\mathbf{\Delta} \mathbf{A})} - \mathbf{I}) \cdot \mathbf{\Delta} \mathbf{B}.
\end{aligned}
\end{equation}
After the discretization, the discretized version of Eq.~\ref{pure_ssm} with step size $\mathbf{\Delta}$ can be rewritten in the following RNN form:
\begin{equation}
\begin{aligned}
\label{eq:discrete_lti}
h_t &= \mathbf{\overline{A}}h_{t-1} + \mathbf{\overline{B}}x_{t}, \\
y_t &= \mathbf{C}h_t + \mathbf{D}x_{t}.
\end{aligned}
\end{equation}
Finally, the model computes output through a global convolution. 
\begin{equation}
\begin{aligned}
\label{eq:conv}
\mathbf{\overline{K}} &= (\mathbf{C}\mathbf{\overline{B}}, \mathbf{C}\mathbf{\overline{A}}\mathbf{\overline{B}}, \dots, \mathbf{C}\mathbf{\overline{A}}^{\mathtt{L}-1}\mathbf{\overline{B}}), \\
\mathbf{y} &= \mathbf{\overline{K}} \circledast \mathbf{x} + \mathbf{D} * \mathbf{x},
\end{aligned}
\end{equation}
where $\mathtt{L}$ is the length of the input sequence $\mathbf{x}$, $\overline{\mathbf{K}} \in \mathbb{R}^{\mathtt{L}}$ is a structured convolutional kernel and $\circledast$ denotes convolution operation. As recent advanced SSM, Mamba~\cite{ref11_MAMBA} proposes S6 to improve 
$\overline{\rm \textbf{B}}$, ${\rm \textbf{C}}$ and $\rm \Delta$ to be input-dependent, thus allowing for a dynamic feature representation.

\subsection{Problem Formulation}
\label{problem_formulation}

\textbf{Classic blind SR:} Generally, the degradation model of an LR image $\mathbf{I^{LR}}$ can be formulated as follows:

\begin{equation}
  \mathbf{I^{LR}} = (\mathbf{I^{HR}}\otimes \mathbf{k}){\downarrow}_s + \mathbf{n},
\end{equation}
where $\mathbf{I^{HR}}$ is the HR image, $\mathbf{k}$ is a blur kernel, $\otimes$ denotes convolution operation, ${\downarrow}_s$ is downsampling operation controlled by scale factor $s$ and $\mathbf{n}$ refers to Gaussian noise. Under blind settings, image SR aims at super-resolving input LR images without knowing the true degradation information. 

\textbf{Real-world SR:} As a variant of classic blind SR, real-world SR \cite{ref1_realesrgan,ref2_bsrgan} adopts more intricate degradation procedures. Specifically, This degradation setting introduces comprehensive degradation operations (\emph{e.g.}, blur, noise, down-sampling, and JPEG compression) and controls the degree of each operation by randomly sampling the respective hyper-parameters. Moreover, random shuffle of degradation orders and second-order degradation are applied to increase degradation complexity.

\subsection{Proposed Framework}

Our framework consists of an encoder, a degrader, and a generator. 
As illustrated in Fig.~\ref{Framework}, our framework can be divided in to two stages (\emph{i.e.}, LR image generation and HR image reconstruction). During the training phase, the encoder and the degrader are employed to extract discriminative representations from LR images. Meanwhile, the generator incorporates the degradation representation to super-resolve the LR image. During the inference, only the encoder and the generator are employed to produce the SR result.

\begin{figure*}[t]
		\centering
		\includegraphics[width=1.0\linewidth]{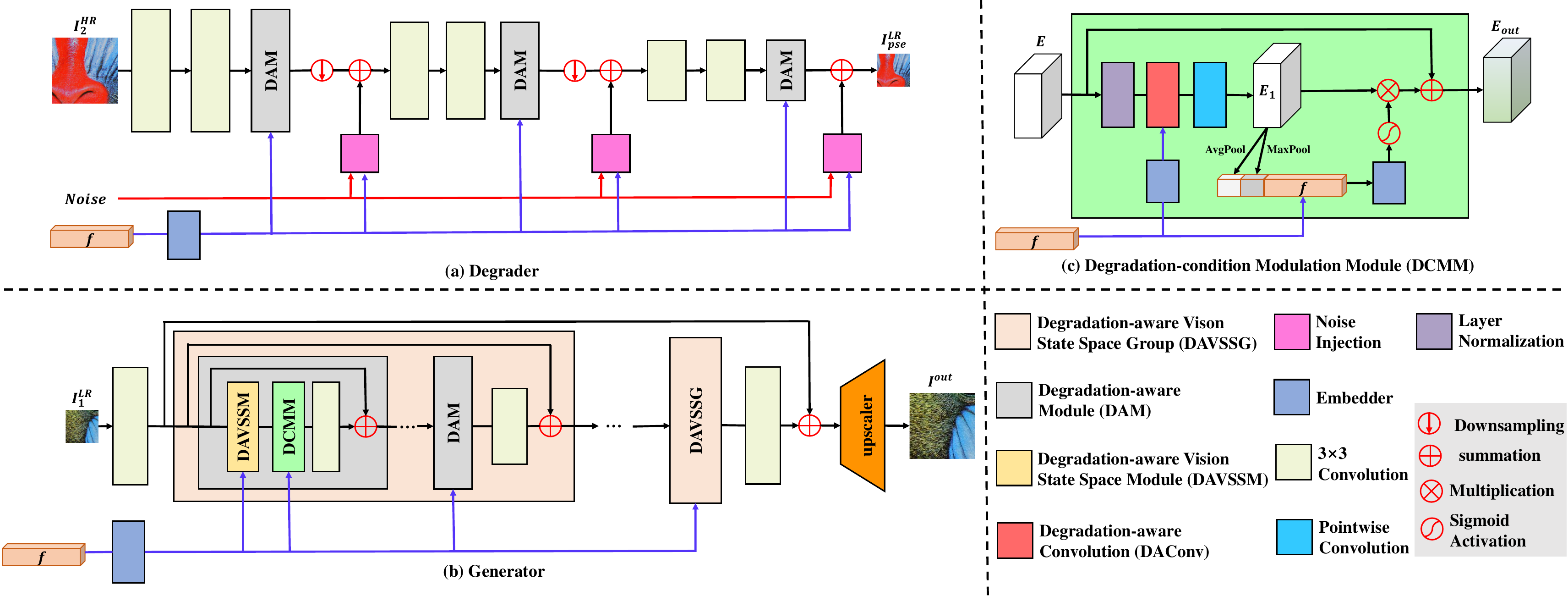}
		\caption{Network architecture of our degrader (a), generator (b) and DCMM (c).}
		\label{MainNet}
\end{figure*}

\begin{figure*}[t]
		\centering
		\includegraphics[width=1.0\linewidth]{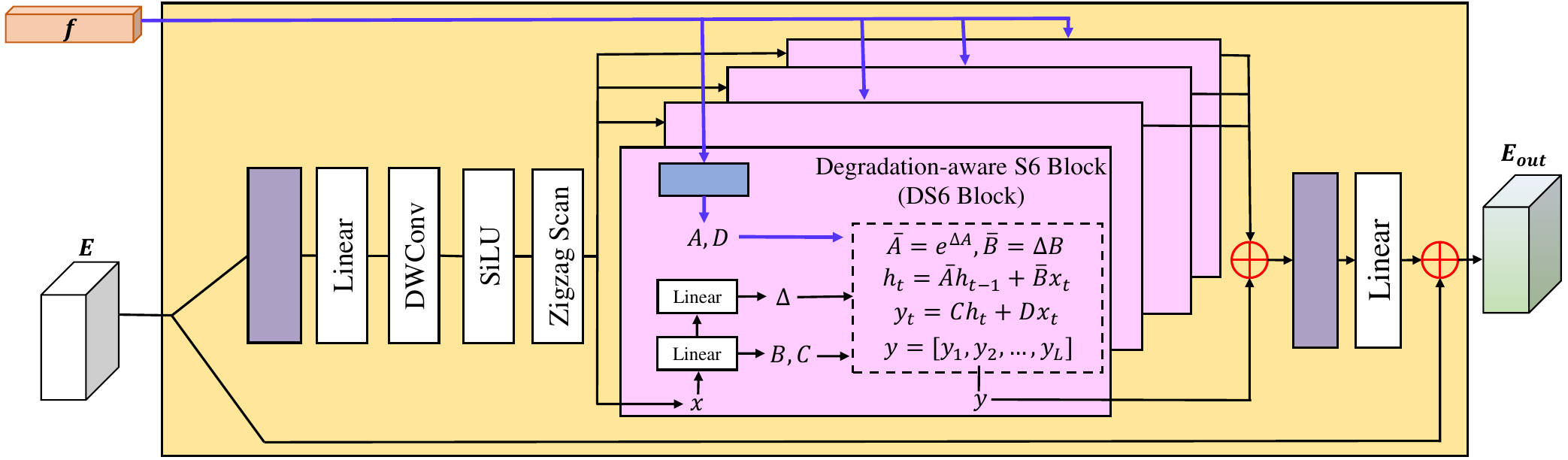}
		\caption{Network architecture of Degradation-aware Vision State Space Module (DAVSSM).}
		\label{DAVSSM}
\end{figure*}

\subsection{Degradation Representation Learning}

Degradation representation learning aims to extract implicit degradation information from LR images in a self-supervised manner.

\subsubsection{Image Re-degradation}

First, an LR image $\mathbf{I^{LR}_{1}}\in\mathbb{R}^{\mathtt{H}\times\mathtt{W}\times3}$ is fed to an encoder to obtain the degradation representation $\mathbf{f}\in \mathbb{R}^{\mathtt{C_f}}$:
\begin{equation}
\mathbf{f}={\rm Encoder}(\mathbf{I^{LR}_{1}}),
\end{equation}
where $\mathtt{C}$ is the number of degradation representation dimension. {Following the protocol in \cite{ref3_dasr}, the encoder consists of six $3\times3$ convolution layers at three different resolution levels. After the last convolution layer, an average pooling layer is adopted to obtain $\mathbf{f}$.}

Then, the degrader takes the extracted degradation representation $\mathbf{f}$ and an HR image $\mathbf{I^{HR}_{2}} \in \mathbb{R}^{s\mathtt{H}\times{s}\mathtt{W}\times3}$ as input to reproduce the corresponding pseudo LR image $ \mathbf{{I}_{pse}^{LR}} \in \mathbb{R}^{\mathtt{H}\times\mathtt{W}\times3}$:
\begin{equation}
{\mathbf{{I}_{pse}^{LR}}}={\rm Degrader}(\mathbf{I_{2}^{HR}},\mathbf{f}),
\end{equation}
To synthesize realistic LR images with diverse degradations, our degrader progressively simulates degradation injection in feature maps at different resolution conditioned on the degradation representation $\mathbf{f}$. Specifically,  as illustrated in Fig.~\ref{MainNet}(a), the input HR image $\mathbf{I_2^{HR}}$ is fed to vanilla $3\times3$ convolution layers to obtain shallow features. Meanwhile, input degradation representation $\mathbf{f}$ is passed to a embedder for feature compression. Then conditioned on degradation information, degradation-aware module (DAM, detailed in Sec.\ref{DASR}) and noise injection module are adopted to inject contamination at different resolution levels while degrading the latent feature to an LR image.

\textbf{Content Consistency Loss.} An content loss between the synthesized pseudo LR image and the input LR image is employed for optimization:
\begin{equation}
\mathcal{L}_{con}=||g(\mathbf{I^{LR}_{2}})-g(\mathbf{{I}^{LR}_{pse}})||_1,
\end{equation}
where $g(\cdot)$ is a $3\times3$ Gaussian filter.

\textbf{Adversarial Loss.} We impose an adversarial loss on the synthetic LR image $\mathbf{{I}^{LR}_{pse}}$ to constrain it in the same degradation domain with $\mathbf{I^{LR}_{2}}$. The adversarial losses used for the discriminator and degrader are defined as: 
\begin{align}
\mathcal{L}_{adv}^{Dis}= & \mathbb{E}_{\mathbf{I^{LR}_2}}[{\rm log}(1-{\rm Net}_{Dis}(\mathbf{I^{LR}_2}))] + \notag \\ 
& \mathbb{E}_{\mathbf{I^{LR}_{pse}}}[{\rm log}({\rm Net}_{Dis}(\mathbf{I^{LR}_{pse}}))],
\end{align}
\begin{align}
\mathcal{L}_{adv}^{Degrader}=\mathbb{E}_{\mathbf{I^{LR}_{pse}}}[{\rm log}(1-{\rm Net}_{Dis}(\mathbf{I^{LR}_{pse}}))],
\end{align}
where ${\rm Net}_{Dis}(\cdot)$ represents the output of the discriminator network, which is similar to UDASR~\cite{ref94_udasr}.

\textbf{Degradation Consistency Loss.} In addition to the above content loss and adversarial loss, we further adopt a degradation consistency loss to enforce the synthetic LR $\mathbf{I^{LR}_{pse}}$ image have degradations similar to the target LR image $\mathbf{I^{LR}_{2}}$. We introduce a regression loss between the predicted degradation representations using the L1 distance as follows:
\begin{align}
\label{degradation_consistency_loss}
\mathcal{L}_{consist}=||{\rm Encoder}(\mathbf{I^{LR}_{2}})-{\rm Encoder}(\mathbf{{I}^{LR}_{pse}})||_1.
\end{align}
By encouraging the synthesized LR images to reproduce the diverse degradation details in the input LR images, full degradation information is captured in the degradation representations. Note that, to avoid the encoder to memorize the degradation in $\mathbf{I^{LR}_1}$ rather than learning general degradation information, $\mathbf{I^{LR}_{2}}$ has different contents with $\mathbf{I^{LR}_1}$ but shares the same degradation.

Then the overall loss for re-degradation branch is defined as:
\begin{align}
\label{loss_rd}
\mathcal{L}_{RD}=\mathcal{L}_{con} + 0.01 \times \mathcal{L}_{adv}^{Degrader} + 0.1 \times \mathcal{L}_{consist}
\end{align}

\subsubsection{Degradation Representation Distribution Alignment}

To constrain the learned degradation representations in a bounded space for better generalization performance, we propose to embed the representations into a specific target distribution (as illustrated in Fig.~\ref{distribution_fig}(b)) during training. In our method, an energy distance is introduced to measure the distance between degradation representation space and the target distribution space. Specifically, $b$ LR images are first randomly selected and encoded into $\left\{\mathbf{f}_1, \mathbf{f}_2 ...\mathbf{f}_b\right\}$ using our encoder, where $\mathbf{f}_i\in{\mathbb{R}}^{{\mathtt{C_f}}}$ is degradation representation of the $i^{th}$ image. Then, we sample $m$ samples from a pre-defined distribution (\emph{e.g.}, Gaussian distribution), obtaining $\left\{\mathbf{t}_1, \mathbf{t}_2 ...\mathbf{t}_m\right\}$. Here, $\mathbf{t_j}\in{\mathbb{R}}^{{\mathtt{C}}}$ is the $j^{\rm th}$ sample. Next, by minimizing energy distance, the representation distribution alignment loss is formulated as:
{\begin{align}
\label{distribution_LOSS}
  \mathcal{L}_{distri}=&\frac{2}{bm}\sum_{i=1}^b\sum_{j=1}^m{\Vert{\mathbf{f}_i-\mathbf{t}_j}\Vert}_2
  \notag \\
  -&\frac{1}{b^2}\sum_{i=1}^b\sum_{j=1}^b{\Vert{\mathbf{f}_i-\mathbf{f}_j}\Vert}_2
  \notag \\
  -&\frac{1}{m^2}\sum_{i=1}^m\sum_{j=1}^m{\Vert{\mathbf{t}_i-\mathbf{t}_j}\Vert}_2.
\end{align}}

\subsubsection{Discussion}

Previous methods commonly use contrastive learning to extract discriminative representations by distinguishing different degradations~\cite{ref3_dasr}. As a result, these representations focus on the degradation differences in a batch while overlooking their common components. Since a finite batch size is not able to cover the whole degradation space, the learned representations cannot well capture subtle degradation difference. Besides, degradation representations generated by previous methods~\cite{ref3_dasr,ref5_kdsr,ref6_mrda} collapse into a subspace in a certain distribution (\emph{i.e.}, Fig.~\ref{distribution_fig} (a)). This indicates that subtle degradation differences cannot be distinguished. In contrast, representations learned by our method are expected to preserve full degradation details by reconstructing the input LR images, thereby obtaining more accurate degradation information. Moreover, with our degradation representation distribution alignment loss, the representations are embedded into a regular pre-defined space, which helps decouple different degradations and amplifies their differences between them. Then differences between various degradations can be well recognized to generate more discriminative and robust representations (as shown in Fig.~\ref{distribution_fig} (b)).

\subsection{Degradation-aware SR}
\label{DASR}

\begin{figure}[t]
		\centering
		\includegraphics[width=1.0\linewidth]{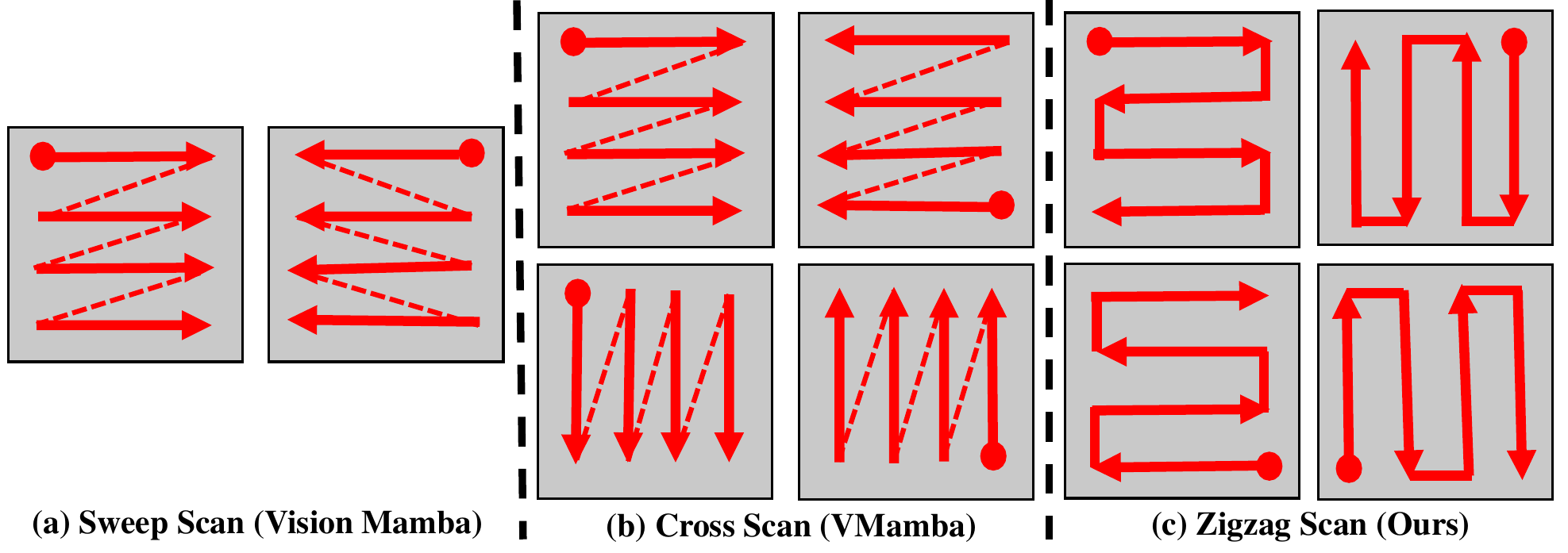}
		\caption{An illustration of different scan methods in visual backbones based on Mamba model (\emph{e.g.}, Vision Mamba~\cite{ref8_zhuvisionmamba} and VMamba~\cite{ref7_liuvmamba}).}
		\label{Scan_methods}
\end{figure}

\subsubsection{Overview}

With degradation information being encoded into the representations, degradation-aware super-resolution is performed to super-resolve the input LR image conditioned on the degradation information, as illustrated in Fig.~\ref{MainNet} (b). Specifically, the LR image $\mathbf{I_1^{LR}}$ is first fed to a convolution layer to obtain shallow feature. Then the shallow feature is passed to Degradation-aware Vision State Space Groups (DAVSSGs) to obtain deep feature, and each DAVSSG contains several Degradation-aware Modules (DAMs). Besides, a convolution layer is implemented at the end of DAVSSG to refine features extracted from DAMs. For each DAM, a Degradation-aware Vision State Space Module (DAVSSM) and a Degradation-condition Modulation Module (DCMM) are designed to build long-range dependencies and enhance local pixel based on learned degradation representations, respectively. Finally, The reconstruct feature is fed to a upscaler to obtain super-resolved image $\mathbf{I^{out}}\in\mathbb{R}^{s\mathtt{H}\times{s}\mathtt{W}\times3}$.

\subsubsection{Degradation-aware Vision State Space Module (DAVSSM)}

The original Mamba Module is designed for the 1-D sequence, which is not suitable for SR tasks requiring spatial-aware understanding. To this end, we introduce the DAVSSM, which incorporates the multi-directional sequence modeling for the vision tasks. Due to the computational efficiency and long-range modeling ability of SS2D block in VMamba~\cite{ref7_liuvmamba}, we also follow the \texttt{Norm} $\rightarrow$ \texttt{Linear} $\rightarrow$ \texttt{DWConv} $\rightarrow$ \texttt{SS2D} $\rightarrow$ \texttt{Norm} $\rightarrow$ \texttt{Linear} flow. Moreover, to achieve flexible degradation-aware adaption, we propose a degradation-pluggable mechanism (as shown in Fig.~\ref{DAVSSM}). Specifically, operations of DAVSSM are presented in Algorithm~\ref{alg:DAVSSM}.

\textbf{Zigzag Scan:} Prior researchs~\cite{ref7_liuvmamba,ref8_zhuvisionmamba} have demonstrated the efficacy of using multiple scanning orders to improve performance (\emph{e.g.}, row-wise and column-wise scans in multiple directions, as shown in Fig.~\ref{Scan_methods}(a)(b)). Previous scanning order can only cover one type of 2D direction (\emph{e.g.}, left to right), which causes spatial discontinuity when moving to a new row or column~\cite{ref9,ref10}. Moreover, as the parameter $\mathbf{\overline{A}}$ in Eq.~\ref{eq:conv} serves as a decaying term, the spatial discontinuity causes abrupt changes in degrees of decaying of adjacent tokens, compounding the semantic discontinuity and resulting in unnatural stylized textures. Inspired by~\cite{ref9,ref12}, which proposes Continuous 2D Scanning for semantic continuity, we implement a Zigzag Scan (as shown in Fig.~\ref{Scan_methods}(c)). The proposed method starts with 4 vertices, with the first clockwise column (or row) as the starting scan-line, aiming at preserving spatial and semantic continuity.

\textbf{Degradation-aware S6 Block (DS6 Block):} Different from $\mathbf{A}$ and $\mathbf{D}$ from a certain concrete embedding space in standard S6 block, we introduce a dynamical weights generation scheme. Specifically, we predict $\mathbf{A}$ and $\mathbf{D}$ from degradation representation $\mathbf{f}$:
\begin{equation}
\mathbf{A}, \mathbf{D} = {\rm Embedder}(\mathbf{f}),
\end{equation}
where $\mathbf{A} \in \mathbb{R}^{\mathtt{E}\times \mathtt{N}}$, $\mathbf{D} \in \mathbb{R}^{\mathtt{E} \times \mathtt{1}}$. $\mathtt{E}$ and $\mathtt{N}$ represent expanded dimension size and SSM dimension, respectively. We design the DS6 block based on 2 aspects. \textbf{(1) Degradation Selectivity:} 
To enable the adaption of hidden state based on degradation information while maintaining degradation selectivity,
we predict $\mathbf{A}$ from degradation embedding space instead of concrete embedding. \textbf{(2) Efficiency:} As shown in Eq.~\ref{eq:conv}, weighting parameters $\mathbf{A}$ and $\mathbf{D}$ are expanded to obtain convolutional kernel and channel-wise scale factor, respectively. The dynamical global convolution kernel maintains efficient computation by parallel operations while adapting to various degradation levels.

\begin{algorithm}[t]
\caption{\name{} Process}
\small
\label{alg:DAVSSM}
\begin{algorithmic}[1]
\REQUIRE{
    shallow feature $\mathbf{E}$: \textcolor{shapecolor}{$(\mathtt{C}, \mathtt{H}, \mathtt{W})$},\\
    degradation representation $\mathbf{f}$: \textcolor{shapecolor}{$(\mathtt{C_f},)$}
}
\ENSURE{reconstruct feature $\mathbf{E_{out}}$: \textcolor{shapecolor}{$(\mathtt{C}, \mathtt{H}, \mathtt{W})$}}

\STATE \textcolor{gray}{\text{/* pre-proces image feature $\mathbf{E}$ */}}

\STATE $\mathbf{E'}$: \textcolor{shapecolor}{$(\mathtt{C}, \mathtt{H}, \mathtt{W})$} $\leftarrow$ $\mathbf{LayerNorm}(\mathbf{E})$

\STATE $\mathbf{E'}$ : \textcolor{shapecolor}{$(\mathtt{E}, \mathtt{H}, \mathtt{W})$} $\leftarrow$ $\mathbf{Linear}(\mathbf{E'})$

\STATE $\mathbf{E'}$ : \textcolor{shapecolor}{$(\mathtt{E}, \mathtt{H}, \mathtt{W})$} $\leftarrow$ $\mathbf{SiLU}(\mathbf{DWConv}(\mathbf{E'}))$

\STATE \textcolor{gray}{\text{/* process with four DS6 Blocks, sequence length $\mathtt{L}=\mathtt{H*W}$ */}}
\FOR{$p$ in \{$path1$, $path2$, $path3$, $path4$\}}
    \STATE $\mathbf{x}_p$: \textcolor{shapecolor}{$(\mathtt{L}, \mathtt{E})$} $\leftarrow$ $p(\mathbf{E'})$
    
    \STATE $\mathbf{B}_p$: \textcolor{shapecolor}{$(\mathtt{L}, \mathtt{N})$} $\leftarrow$ $\mathbf{Linear}^{\mathbf{B}}_p(\mathbf{x}_p)$
    
    \STATE $\mathbf{C}_p$: \textcolor{shapecolor}{$(\mathtt{L}, \mathtt{N})$} $\leftarrow$ $\mathbf{Linear}^{\mathbf{C}}_p(\mathbf{x}_p)$
    
    \STATE \textcolor{gray}{\text{/* softplus ensures positive $\mathbf{\Delta}_p$ */}}
    
    \STATE $\mathbf{\Delta}_p$: \textcolor{shapecolor}{$(\mathtt{L}, \mathtt{E})$} $\leftarrow$ $\log(1 + \exp(\mathbf{Linear}^{\mathbf{\Delta}}_p(\mathbf{x}_p) + \mathbf{Parameter}^{\mathbf{\Delta}}_p))$

    \STATE \textcolor{gray}{\text{/* degradation-aware parameters */}}
    
    \STATE $\mathbf{A}_p$: \textcolor{shapecolor}{$(\mathtt{N}, \mathtt{E})$} $\leftarrow$ \textcolor{blue}{$\mathbf{Embedder}^{\mathbf{A}}_p(\mathbf{f})$}

    \STATE $\mathbf{D}_p$: \textcolor{shapecolor}{$(\mathtt{E},)$} $\leftarrow$ \textcolor{blue}{$\mathbf{Embedder}^{\mathbf{D}}_p(\mathbf{f})$}

    \STATE \textcolor{gray}{\text{/* discretization process */}}

    \STATE $\overline{\mathbf{A}}_p$: \textcolor{shapecolor}{$(\mathtt{L}, \mathtt{N}, \mathtt{E})$} $\leftarrow$ $\exp(\mathbf{\Delta}_p \bigotimes \mathbf{A}_p)$ 
    
    \STATE $\overline{\mathbf{B}}_p$ : \textcolor{shapecolor}{$(\mathtt{L}, \mathtt{N}, \mathtt{E})$} $\leftarrow$ $\mathbf{\Delta}_p \bigotimes \mathbf{B}_p$
    
    \STATE $\mathbf{y}_p$ : \textcolor{shapecolor}{$(\mathtt{L}, \mathtt{E})$} $\leftarrow$ $\mathbf{SSM}(\overline{\mathbf{A}}_p, \overline{\mathbf{B}}_p, \mathbf{C}_p, \mathbf{D}_p)(\mathbf{x}_p)$

    \STATE $\mathbf{y}_p$ : \textcolor{shapecolor}{$(\mathtt{E}, \mathtt{H}, \mathtt{W})$} $\leftarrow$ $\mathbf{Merge}(\mathbf{y}_p)$
    
\ENDFOR

\STATE $\mathbf{E'}$ : \textcolor{shapecolor}{$(\mathtt{E}, \mathtt{H}, \mathtt{W})$} $\leftarrow$ $\mathbf{LayerNorm}(\mathbf{y}_{path1}+\mathbf{y}_{path2}+\mathbf{y}_{path3}+\mathbf{y}_{path4})$

\STATE $\mathbf{E_{out}}$ : \textcolor{shapecolor}{$(\mathtt{C}, \mathtt{H}, \mathtt{W})$} $\leftarrow$ $\mathbf{Linear}(\mathbf{E'}) + \mathbf{E}$

Return: $\mathbf{E_{out}}$ 
\end{algorithmic}
\end{algorithm}

\subsubsection{Degradation-aware Modulation Module (DCMM)} As mentioned in~\cite{ref13_mambair}, since SSMs process flattened feature maps as 1D token sequences, the number of adjacent pixels in the sequence is greatly influenced by the flattening strategy. The over-distance in a 1D token sequence between spatially close pixels can lead to local pixel forgetting (\emph{e.g.}, the patch at row $i$ and column $j$ is no longer adjacent to the patch at row $(i + 1)$ and column $j$ in the row-major scan). Moreover, SSMs lead to notable channel redundancy due a larger number of hidden states to memorize very long-range dependencies. To avoid these problems, we design a DCMM, which contains a Degradation-aware Convolution (DAConv) to compensate for local features and a Degradation-condition Channel Attention (DC$^2$A) to facilitate the expressive power of different channels.

\textbf{Degradation-aware Convolution (DAConv):} To process image feature conditioned on degradation representation, a simple and straightforward scheme is to concatenate degradation representation with image feature and fuse them via convolution operation~\cite{ref15,ref16_udvd}. However, as demonstrated in several recent works~\cite{ref3_dasr,ref14_ikc}, directly convolving the concatenated feature can cause interference since there is a domain gap between these two kinds of representations. Motivated by the observation that convolution kernels of models trained for different restoration levels share similar patterns but have different statistics~\cite{ref3_dasr,ref4,ref17}, we design DAConv whose kernels are dynamically generated according to the input degradation representation. Following the protocol of DASR~\cite{ref3_dasr}, the degradation representation $\mathbf{f}$ is passed to an embedder to produce a convolution kernel $\mathbf{K}\in\mathbb{R}^{\mathtt{C}\times1\times3\times3}$. Then the input image feature $\mathbf{E}$ is processed by layer normalization and depth-wise convolution (using $\mathbf{K}$). And an additional $1\times1$ convolution layer is implemented to produce $\mathbf{E}_1$

\textbf{Degradation-condition Channel Attention (DC$^2$A):} Inspired by CResMD~\cite{ref18_cresmd} which uses controlling variables to rescale different channels to handle multiple degradations, we design a channel attention layer to reweight the output features based on the statistics of both image feature (produced by performing average and max pooling on image feature) and degradation representation. Specifically, the concatenated representation is passed to another embedder and a sigmoid activation layer to generate channel-wise modulation coefficients. Then, the coefficients are used to rescale different channel components in $\mathbf{E}_1$. Finally, the modulated feature is summed up with input feature and fed to the subsequent layers.

\subsubsection{Reconstruction Loss}

As mentioned in Sec.~\ref{problem_formulation}, blind SR methods always follow two blind SR settings: classic blind SR and real-world SR. Under the two settings, the reconstruction loss $\mathcal{L}_{SR}$ is defined by different formulations.

\textbf{Classic blind SR:} Following previous works~\cite{ref3_dasr,ref5_kdsr,ref6_mrda}, classic blind SR networks are commonly trained with manhattan distance.
\begin{align}
\label{loss_sr1}
\mathcal{L}_{SR} = \mathcal{L}_{1} = ||\mathbf{{I}^{HR}_{1}}-\mathbf{{I}^{out}}||_1,
\end{align}

\textbf{Real-world SR:} With more complex degradation conducting in LR image generation, it's difficult to implement a degradation estimator to predict degradation types. Consequently, the perceptual loss $\mathcal{L}_{per}$~\cite{ref19_perloss} and adversarial loss $\mathcal{L}_{adv}^{Generator}$~\cite{ref1_realesrgan} are used to emphasize visual quality. Following the training protocol~\cite{ref1_realesrgan,ref2_bsrgan}, the reconstruction loss is defined as:
\begin{align}
\label{loss_sr2}
\mathcal{L}_{SR} = \mathcal{L}_{1} + \mathcal{L}_{per} + 0.1 \times \mathcal{L}_{adv}^{Generator}
\end{align}

\begin{table*}[t]
 \renewcommand{\arraystretch}{1.2}
 
		\caption{PSNR results achieved on Urban100 for $\times4$ SR. We select 5 typical anisotropic Gaussian kernels (as shown in the upper right corner) and 3 noise levels (\emph{i.e.}, 5, 15 and 25) for evaluation. Results averaged on different noise levels are reported for each blur kernel.}
  \vspace{-10pt}
		\label{tab_ablation1}
		\begin{center}
            \resizebox{1\hsize}{!}{

\begin{threeparttable}[b]

				\begin{tabular}{|l|c|ccc|cccc|ccccc|}
					\hline
					\multirow{3}{*}{Model} & \textbf{Encoder} & \multicolumn{3}{c|}{\textbf{Degrader}} 
                    & \multicolumn{4}{c|}{\textbf{Generator}}
					& \multicolumn{5}{c|}{Blur Kernel\footnotemark[1]} \\
                     \cline{2-9}
                    
					& \multirow{2}{*}{$\mathcal{L}_{distri}$ (Eq.~\ref{distribution_LOSS})}
                    & \multirow{2}{*}{$\mathcal{L}_{RD}$ (Eq.~\ref{loss_rd})}
                    & \multirow{2}{*}{\tabincell{c}{Noise \\ Injection}}
                    & \multirow{2}{*}{DAM} 
                    & \multicolumn{2}{c|}{DAVSSM}
                    & \multicolumn{2}{c|}{DCMM} 
                    & \multirow{2}{*}{\begin{minipage}[b]{0.07\columnwidth}
						\centering
						\raisebox{-.5\height}
      {\includegraphics[width=18pt]{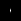}}
					\end{minipage}}
                    & \multirow{2}{*}{\begin{minipage}[b]{0.07\columnwidth}
						\centering
						\raisebox{-.5\height}
      {\includegraphics[width=18pt]{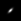}}
					\end{minipage}}
                    & \multirow{2}{*}{\begin{minipage}[b]{0.07\columnwidth}
						\centering
						\raisebox{-.5\height}
      {\includegraphics[width=18pt]{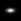}}
					\end{minipage}}
                    & \multirow{2}{*}{\begin{minipage}[b]{0.07\columnwidth}
						\centering
						\raisebox{-.5\height}
      {\includegraphics[width=18pt]{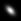}}
					\end{minipage}}
                    & \multirow{2}{*}{\begin{minipage}[b]{0.07\columnwidth}
						\centering
						\raisebox{-.5\height}
      {\includegraphics[width=18pt]{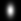}}
					\end{minipage}}
                    \\
                    \cline{6-9}

                   & & &&& Zigzag & \multicolumn{1}{c|}{DS6} & DAConv & DC$^2$A & &  &  & &
                    \tabularnewline			
					\hline
                    \hline
                    
					E1 & \textcolor[RGB]{202,12,22}{\faTimes} &  \faCheck  & \faCheck  & \faCheck  & \faCheck & \faCheck & \faCheck & \faCheck
                      & 24.09 & 23.84 & 23.37 & 22.82 & 22.19
					\tabularnewline
                    \hline
                    \hline
     
					D1 & \faCheck & \textcolor[RGB]{202,12,22}{\faTimes} & \textcolor[RGB]{202,12,22}{\faTimes}  & \textcolor[RGB]{202,12,22}{\faTimes}  & \faCheck  & \faCheck  & \faCheck & \faCheck
                        & 24.11 & 23.80 & 23.27 & 22.70 & 22.09
					\tabularnewline
     
					D2 & \faCheck  & \faCheck & \textcolor[RGB]{202,12,22}{\faTimes}  & \textcolor[RGB]{202,12,22}{\faTimes}   & \faCheck  & \faCheck  & \faCheck & \faCheck
                        & 24.23 & 23.95 & 23.38 & 23.82 & 22.20
                      
					\tabularnewline
     
					D3   & \faCheck & \faCheck & \faCheck  & \textcolor[RGB]{202,12,22}{\faTimes}  & \faCheck  & \faCheck  & \faCheck  & \faCheck
                        & 24.64 & 24.35 & 23.81 & 23.19 & 22.58
					\tabularnewline
                    \hline
                    \hline
                    G1   & \faCheck & \faCheck & \faCheck  & \faCheck  & \textcolor[RGB]{202,12,22}{\faTimes} & \textcolor[RGB]{202,12,22}{\faTimes} & \faCheck  & \faCheck 
                      & 24.49 & 24.28 & 23.72 & 23.10 & 22.41
					\tabularnewline
                    G2   & \faCheck & \faCheck & \faCheck  & \faCheck  & \faCheck  & \textcolor[RGB]{202,12,22}{\faTimes}  & \faCheck & \faCheck
                      & 24.60 & 24.39 & 23.82 & 23.22 & 22.53
					\tabularnewline
                    G3   & \faCheck & \faCheck & \faCheck  & \faCheck  & \faCheck & \faCheck  & \textcolor[RGB]{202,12,22}{\faTimes}  & \textcolor[RGB]{202,12,22}{\faTimes} 
                      & 24.67 & 24.44 & 23.90 & 23.25 & 22.60
					\tabularnewline
                    G4   & \faCheck & \faCheck & \faCheck  & \faCheck  & \faCheck & \faCheck & \faCheck & \textcolor[RGB]{202,12,22}{\faTimes}
                      & 24.81 & 24.56 & 24.03 & 23.37 & 22.72
					\tabularnewline

                    \hline
                    \hline

                    Baseline (Ours) & \faCheck & \faCheck & \faCheck & \faCheck  & \faCheck  & \faCheck  & \faCheck & \faCheck 
                    & 24.92 & 24.65 & 24.13 & 23.51 & 22.83
					\tabularnewline
					\hline	
			\end{tabular}

               \begin{tablenotes}
     \item[1] Parameters ($\lambda_1$, $\lambda_2$, $\theta$) of 5 anisotropic Gaussian kernels: $(0.2, 1.0, 0^{\circ})$, $(0.8, 1.8, 45^{\circ})$, $(1.4, 2.6, 90^{\circ})$, $(2.0, 3.4, 135^{\circ})$, $(2.6, 4.2, 180^{\circ})$.
   \end{tablenotes}
  \end{threeparttable}
            }
		\end{center}
  \vspace{-10pt}
	\end{table*}

\begin{table}[t]
 \renewcommand{\arraystretch}{1.2}
 \vspace{-9pt}
		\caption{Ablation study of distribution alignment loss $\mathcal{L}_{distri}$ (Eq.~\ref{distribution_LOSS}). LR images are generated by the degradations in Table~\ref{tab_ablation1}.}
  \vspace{-10pt}
		\label{tab_ablation_distribution}
		\begin{center}
            \resizebox{1\hsize}{!}{
				\begin{tabular}{|l|c|c|cccc|}
					\hline
					\multirow{2}{*}{Model} & \multirow{2}{*}{\tabincell{c}{Target \\ Distribution}} & \multirow{2}{*}{\tabincell{c}{Distance \\ Metric}}
                    & \multicolumn{4}{c|}{Datasets} 
					\tabularnewline

                   & &  & Set5 & Set14 & B100 & Manga109
                    \tabularnewline			
					\hline
                    \hline
                   TD1 & Exponential & ED & 28.12 & 25.91 & 25.37 & 26.33
                     \tabularnewline	
                    TD2 & Uniform & ED & 28.07 & 25.85 & 25.31 & 26.31
                     \tabularnewline	
                    TD3 & Chi-square & ED & 28.11 & 25.88 & 25.33 & 26.29
                     \tabularnewline	

                     \hline
                    \textbf{Baseline} & Gaussian & ED & \textbf{28.16} & \textbf{25.97} & \textbf{25.41} & \textbf{26.41}
					\tabularnewline
                    \hline

                    DM1 & Gaussian & KLD & 27.69 & 25.71 & 25.29 & 25.84
                     \tabularnewline	
                    DM2 & Gaussian & JSD & 27.71 & 25.72 & 25.29 & 25.88
                     \tabularnewline	
                    DM3 & Gaussian & HD & 27.82 & 25.80 & 25.33 & 26.04
                     \tabularnewline	
                    
					\hline	
			\end{tabular}}
		\end{center}
  \vspace{-10pt}
	\end{table}

\begin{figure}[t]
		\centering
        \vspace{-9pt}
		\includegraphics[width=1.0\linewidth]{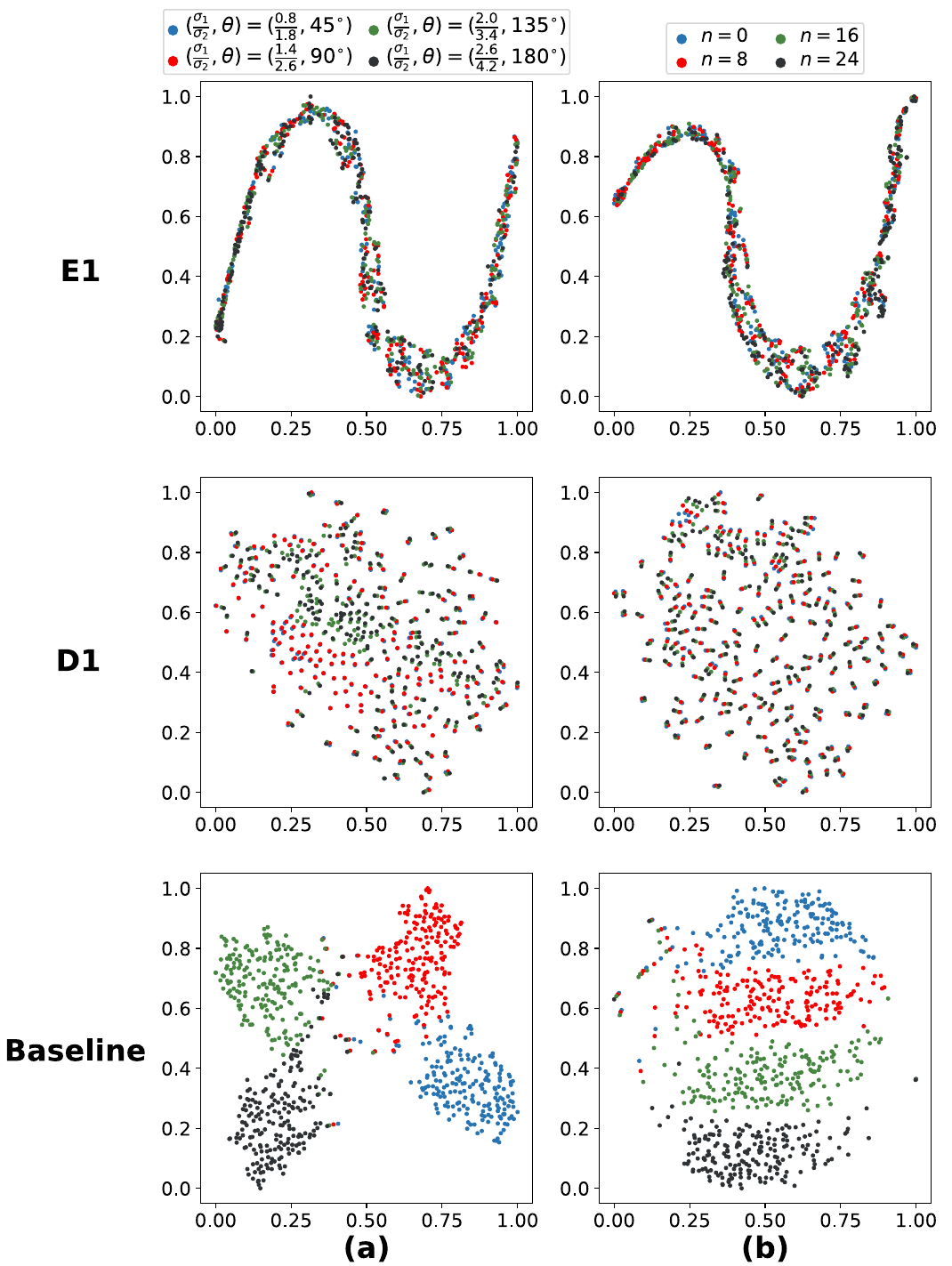}
		\caption{Visualization of representations for different model variants on degradations with (a) different blur kernels and (b) noise levels.}
		\label{scatter1}
\end{figure}

\begin{figure*}[t]
		\centering
		\includegraphics[width=1.0\linewidth]{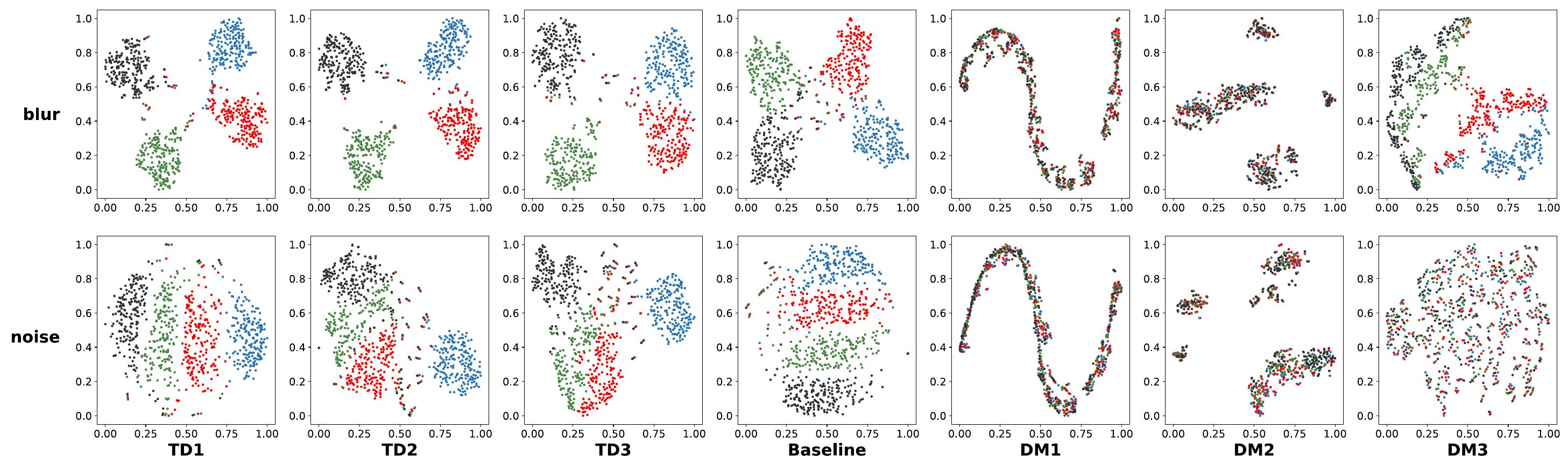}
		\caption{Visualization of representations for different target distributions and distance metrics. Note that, the representations are extracted from degradations in Fig.~\ref{scatter1}.}
		\label{scatter2}
\end{figure*}

\section{Experiments}

\subsection{Implementation Details}
\label{imple_details}

Following the protocol in~\cite{ref14_ikc,ref3_dasr}, we use 800 images in DIV2K~\cite{ref20_div2k} and 2650 images in Flickr2K~\cite{ref21_flickr2k} as the training set, and include 5 benchmark datasets (Set5~\cite{ref23_set5}, Set14~\cite{ref24_set14}, B100~\cite{ref25_B100} and Urban100~\cite{ref26_Urban100} and Manga109~\cite{ref22_manga109}) for evaluation. The kernel size is fixed to $21\times21$. We first trained our method on isotropic Gaussian kernels. The ranges of kernel width $\sigma$ were set to $[0.2,4.0]$ for $\times4$ SR. Then, our framework was trained on more general degradations. Specifically, anisotropic Gaussian kernels with kernel width $\sigma_1,\sigma_2{\sim}U(0.2,4)$ and rotation angle $\theta{\sim}U(0,\pi)$ are employed. In addition, the range of noise level is set to $[0, 25]$. Finally, we conducted experiments on real-world degradation~\cite{ref1_realesrgan,ref2_bsrgan}. Following Real-ESRGAN~\cite{ref1_realesrgan}, we add 10324 images in OutdoorSceneTraining~\cite{ref28_outdoortraining} dataset to training set. And we generate LR images synthesized by the second-order degradation generation scheme~\cite{ref1_realesrgan}. The size of LR patch is set to $64\times64$ for all training settings.

During training, 16 HRs were randomly selected. Then, we randomly selected 16 degradation models from the above ranges to generate LR images. Next, 32 HR-LR patch pairs (2 patch pairs from each image, as illustrated in Fig.~\ref{Framework}) were divided into 2 groups. One group was fed to the degrader while another one was fed to the generator and the encoder. Sample number $m$ in $\mathcal{L}_{distri}$ (Eq.~\ref{distribution_LOSS}) was set to 64. In Algorithm~\ref{alg:DAVSSM}, image feature channel number $\mathtt{C}$, expanded dimension size $\mathtt{E}$ and SSM dimension $\mathtt{N}$ are set to 64, 128 and 16, respectively. We adopted the Adam optimizer~\cite{ref27_adam} with the momentum of $\beta_1=0.9$, $\beta_2=0.999$ for optimization. We trained the whole network for 1000 epochs. The initial learning rate was set to $0.0001$ and decreased to half after every 200 epochs. The overall loss $\mathcal{L}_{overall}$ consists of three components: re-degradation loss $\mathcal{L}_{RD}$ (Eq.~\ref{loss_rd}), degradation representation distribution alignment loss $\mathcal{L}_{distri}$ (Eq.~\ref{distribution_LOSS}) and reconstruction loss $\mathcal{L}_{SR}$ (Eq.~\ref{loss_sr1} or Eq.~\ref{loss_sr2}). And $\mathcal{L}_{overall}$ is defined as:
\begin{align}
\mathcal{L}_{overall} = \mathcal{L}_{SR} + \mathcal{L}_{RD} + {\lambda} \times \mathcal{L}_{distri},
\end{align}
where ${\lambda}=0.01$.

\begin{figure}[t]
		\centering
		\includegraphics[width=1.0\linewidth]{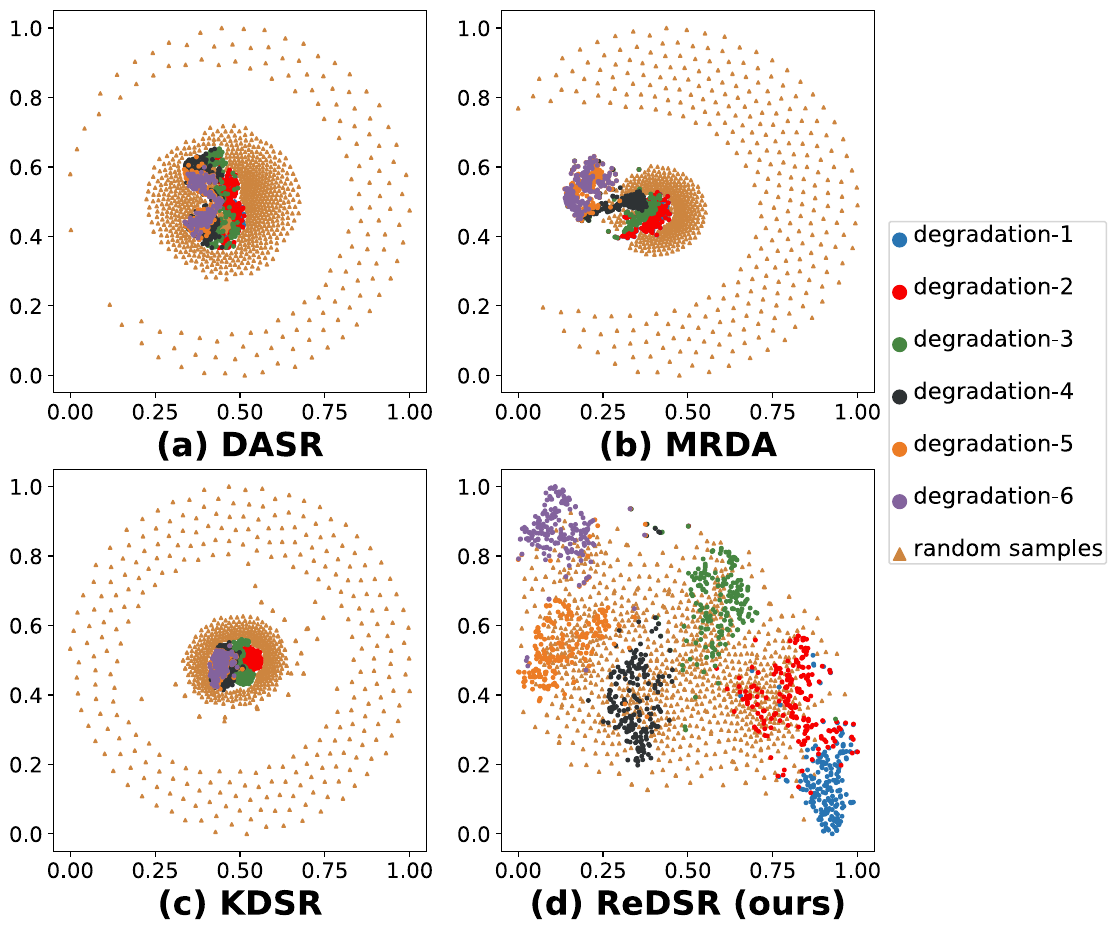}
		\caption{Visualization of representations extracted from LR images and pre-defined distribution (\emph{i.e.}, Gaussian distribution).}
		\label{scatter3}
\end{figure}

\subsection{Model Analysis}

We conduct ablation study on general degradation with anisotropic Gaussian kernels and noises.


\begin{table*}[t]
		\caption{PSNR and SSIM results achieved on Gaussian8 kernels for $\times4$ SR. Methods marked with $^*$ are degradation estimation based approaches, while other methods are degradation representation learning based approaches. Best and second best performance are in \textcolor[RGB]{202,12,22}{red} and \textcolor{blue}{blue} colors, respectively. Running time is averaged on Set14.}
  \vspace{-10pt}
        \renewcommand{\arraystretch}{1.1} 
		\label{tab1}
		\begin{center}
			\small
            \resizebox{1\hsize}{!}{
                \scriptsize
				\begin{tabular}{|lcc|cc|cc|cc|cc|cc|}
					\hline 
					 &  &
     
					& \multicolumn{2}{c|}{\multirow{1}{*}{Set5}} 
					& \multicolumn{2}{c|}{\multirow{1}{*}{Set14}} 
                    & \multicolumn{2}{c|}{\multirow{1}{*}{B100}}   & \multicolumn{2}{c|}{\multirow{1}{*}{Urban100}} 
                    & \multicolumn{2}{c|}{\multirow{1}{*}{Manga109}} 
					\tabularnewline
					
					\multicolumn{1}{|l}{Methods} & Param (M) & Time (ms)
					& PSNR & SSIM 
					& PSNR & SSIM 
					& PSNR & SSIM 
					& PSNR & SSIM 
                    & PSNR & SSIM 
					\tabularnewline
					\hline
					
					MANet${^*}$~\cite{ref38_MANet} 
                        & 9.90 & 102
					& 30.43 & 0.8213 
					& 27.40 & 0.7464
                        & 26.63 & 0.7231
					& 24.88 & 0.7485
                        & 29.90 & 0.8711
					\tabularnewline

     DARSR${^*}$~\cite{ref83_DARSR} 
                        & 3.54 & 62160
					& 28.45 & 0.7877
					& 26.17 & 0.7143
                        & 25.20 & 0.6920
					& 24.09 & 0.7135
                        & 28.84 & 0.8539
					\tabularnewline

                        DANv1${^*}$~\cite{ref37_DAN}
					& 4.33 & 151
					& 31.89 & 0.8864
					& 28.42 &  0.7687
                        & 27.51 & 0.7248
					& 25.86 &  0.7721
                    & 30.50 & 0.9037
					\tabularnewline

            DANv2${^*}$~\cite{ref81_DANv2}
					& 4.71 & 152
					& 32.00 & 0.8885
					& 28.50 & 0.7715
                        & 27.56 & 0.7277
					& 25.94 & 0.7748
                    & 30.45 & 0.9037
					\tabularnewline

           DCLS${^*}$~\cite{ref82_DCLS}
					& 13.63 & 133
					& \textcolor{blue}{32.12} & 0.8890
					& \textcolor{blue}{28.54} & \textcolor{blue}{0.7728}
                        & \textcolor{blue}{27.60} & \textcolor{blue}{0.7285}
					& \textcolor{blue}{26.15} & \textcolor{blue}{0.7809}
                    & \textcolor{blue}{30.86} & \textcolor{blue}{0.9086}
					\tabularnewline

     \hline
     
					DASR~\cite{ref3_dasr}
					& 5.84 & 49
					& 31.46 & 0.8789
					& 28.11 & 0.7603
                        & 27.44 & 0.7214
					& 25.36 & 0.7506
                    & 29.39 & 0.8861
					\tabularnewline

                        CDSR~\cite{ref51_CDSR}
					& 13.23 & 113
					& 31.33 & 0.8328
					& 27.90 & 0.7477
                        & 27.13 & 0.7046
					& 25.25 & 0.7492
                        & 29.18 & 0.8642
					\tabularnewline

                        CMDSR~\cite{ref80_CMDSR}
					& 1.48 & 40
					& 29.10 & 0.8146
					& 26.57 & 0.7239
                        & 26.19 & 0.6980
					& 23.67 & 0.7211
                    & 27.92 & 0.8487
					\tabularnewline

                        MRDA~\cite{ref6_mrda}
					& 5.84 & 57
					& 31.98 & 0.8872
					& 28.42 & 0.7671
                        & 27.55 & 0.7254
					& 25.90 & 0.7734
                    & 30.51 & 0.9088
					\tabularnewline

                        KDSR~\cite{ref5_kdsr}
					& 5.80 & 63
					& 32.02 & \textcolor{blue}{0.8892}
					& 28.46 & 0.7761
                        & 27.52 & 0.7281
					& 25.96 & 0.7760
                    & 30.58 &  0.9026
					\tabularnewline

					ReDSR (Ours) 
					& 5.79 & 43
					&\textcolor[RGB]{202,12,22}{32.25} & \textcolor[RGB]{202,12,22}{0.9028}
					& \textcolor[RGB]{202,12,22}{28.70} & \textcolor[RGB]{202,12,22}{0.7892}
                    & \textcolor[RGB]{202,12,22}{27.72} & \textcolor[RGB]{202,12,22}{0.7381}
					& \textcolor[RGB]{202,12,22}{26.39} & \textcolor[RGB]{202,12,22}{0.8065}
                    & \textcolor[RGB]{202,12,22}{31.07} & \textcolor[RGB]{202,12,22}{0.9142}
					\tabularnewline

					\hline	
			\end{tabular}
   }
		\end{center}
  \label{comparisonwsotas_iso}
	\end{table*}

\begin{figure*}[t]
		\centering
         \vspace{-9pt}
		\includegraphics[width=1\linewidth]{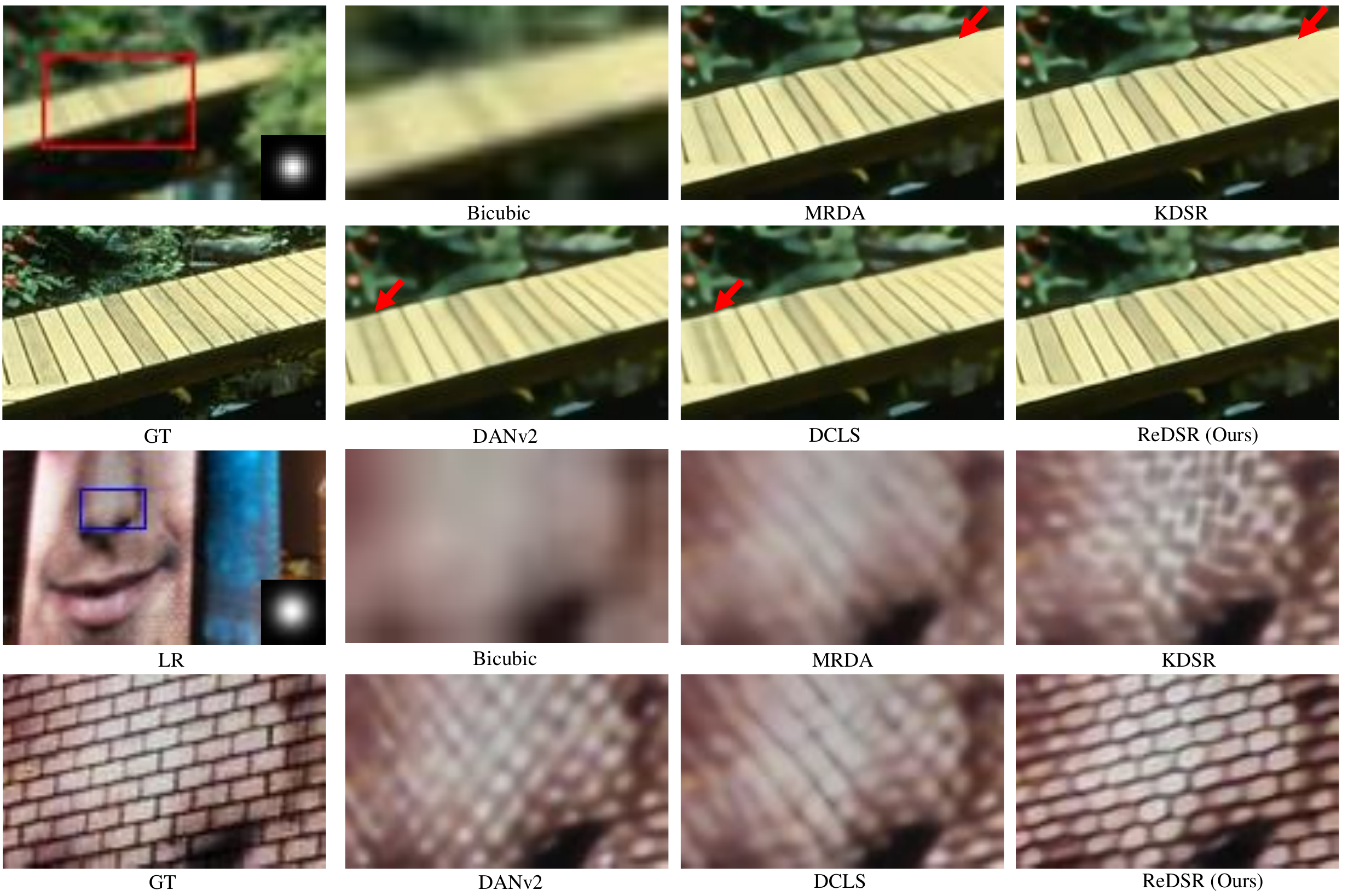}
  \vspace{-9pt}
		\caption{Visual comparison achieved on B100 and Urban100.}
		\label{visual_comparison_iso}
   \vspace{-15.5pt}
\end{figure*}

\begin{figure}[!t]
		\centering
		\includegraphics[width=0.95\linewidth]{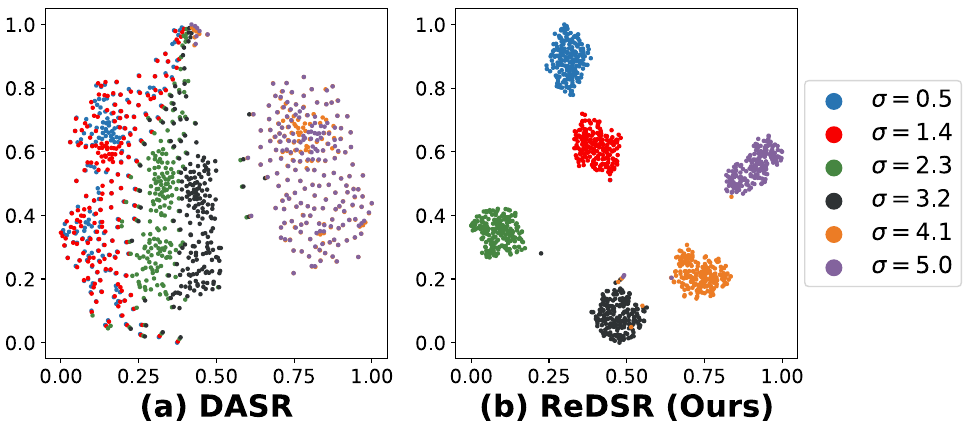}
		\caption{Visualization of representations extracted from LR images with different kernel widths $\sigma$.}
		\label{scatter4_iso}
   \vspace{-20pt}
\end{figure}

\subsubsection{Degradation Representation Learning}
\ 
\newline
\indent \textbf{(1) Encoder}

\textbf{Distribution Alignment Loss $\mathcal{L}_{distri}$:} To promote the learning of degradation information, distribution alignment loss $\mathcal{L}_{distri}$ is employed in our method. To demonstrate its contribution to the final performance, we remove $\mathcal{L}_{distri}$ from our baseline to obtain the model variant E1 for comparison. As we can see in Table~\ref{tab_ablation1}, our baseline surpasses E1 with notable margins. We further visualize the degradation representations extracted from images with various degradations using the t-SNE method~\cite{ref87_tsne} in Fig.~\ref{scatter1}. It can be observed that our baseline can well distinguish different degradations and gather them into discriminative clusters. We further conduct experiments to study the effect of different distribution types (\emph{i.e.}, Gaussian, exponential, uniform, and chi-square distributions). It can be observed from Fig.~\ref{scatter2} that the representations learned by TD1-3 are also discriminative, which further demonstrates the effectiveness of the energy distance. The quantitative results are shown in Table~\ref{tab_ablation_distribution}. It can be observed that our baseline produces comparable performance. As a result, Gaussian distribution is used as the default setting for our method.

The energy distance is used to associate the representation distribution with a pre-defined distribution. As alternatives, Kullback-Leibler divergence (KLD)~\cite{ref67_KLdivergence}, Jensen-Shannon Divergence (JSD)~\cite{ref84_JSdivergence1,ref85_JSdivergence2} and Hellinger distance (HD)~\cite{ref29_Hellingerdistance1,ref30_Hellingerdistance2,ref31_Encyclopedia} are also capable of mapping image space to a pre-defined distribution. To demonstrate the superiority of the energy distance, we replace it with other distance metrics. As shown in Table~\ref{tab_ablation_distribution}, our baseline produces significant performance improvements on different benchmarks. We also visualize the representations extracted from model variants DM1-3 in Fig.~\ref{scatter2}. It can be observed that KLD and JSD cannot generate discriminative clusters (\emph{i.e.}, 5th, and 6th columns in Fig.~\ref{scatter2}). And HD cannot distinguish different noise levels. In contrast, the representations learned by our method gather into several distinct clusters, which further demonstrates that accurate degradation information can be learned by our method.

\begin{table*}[t]
 \tiny
 \caption{PSNR results achieved on Urban100 for $\times4$ SR, testing on Anisotropic Gaussian blur and noise.}
 \vspace{-10pt}
        \renewcommand{\arraystretch}{1.1}
		\label{tab2}
		\begin{center}
			\footnotesize
            \resizebox{1\hsize}{!}{
                    \scriptsize
				\begin{tabular}{|l|l|c|c|ccccccccc|}
    
					\hline 
					\multirow{2}{*}{\tabincell{c}{\\Method}}
					& \multirow{2}{*}{\tabincell{c}{\\\#Params.}}
					& \multirow{2}{*}{\tabincell{c}{\\Time}}
					& \multirow{2}{*}{\tabincell{c}{\\Noise}}
					& \multicolumn{9}{c|}{Blur Kernel}
					\tabularnewline
     
					& & & 
					& \begin{minipage}[b]{0.07\columnwidth}
						\centering
						\raisebox{-.5\height}
      {\includegraphics[width=15pt]{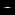}}
					\end{minipage}
     
					& \begin{minipage}[b]{0.07\columnwidth}
						\centering
						\raisebox{-.5\height}{\includegraphics[width=15pt]{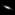}}
					\end{minipage}
					& \begin{minipage}[b]{0.07\columnwidth}
						\centering
						\raisebox{-.5\height}{\includegraphics[width=15pt]{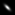}}
					\end{minipage}
					& \begin{minipage}[b]{0.07\columnwidth}
						\centering
						\raisebox{-.5\height}{\includegraphics[width=15pt]{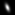}}
					\end{minipage}
					& \begin{minipage}[b]{0.07\columnwidth}
						\centering
						\raisebox{-.5\height}{\includegraphics[width=15pt]{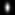}}
					\end{minipage}
					& \begin{minipage}[b]{0.07\columnwidth}
						\centering
						\raisebox{-.5\height}{\includegraphics[width=15pt]{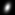}}
					\end{minipage}
					& \begin{minipage}[b]{0.07\columnwidth}
						\centering
						\raisebox{-.5\height}{\includegraphics[width=15pt]{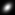}}
					\end{minipage}
					& \begin{minipage}[b]{0.07\columnwidth}
						\centering
						\raisebox{-.5\height}{\includegraphics[width=15pt]{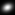}}
					\end{minipage}
					& \begin{minipage}[b]{0.07\columnwidth}
						\centering
						\raisebox{-.5\height}{\includegraphics[width=15pt]{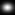}}
					\end{minipage}

					\tabularnewline
					\hline

\multirow{3}{*}{\tabincell{l}
     {DnCNN~\cite{ref86_DnCNN}\\+DANv2~\cite{ref81_DANv2}}}
					& \multirow{3}{*}{\tabincell{l}
     {650K\\+4.71M}} & \multirow{3}{*}{155ms} & 0 
                    & 25.57 & 25.45 & 25.40 & 25.27 & 25.17 & 25.22 & 25.03 & 24.90 & 24.70
					\tabularnewline
					&  & & 10
                        & 24.41 & 24.26 & 24.15 & 23.95 & 23.80 & 23.75 & 23.52 & 23.32 & 23.05
					\tabularnewline
					&  & & 20 
                        & 23.65 & 23.49 & 23.37 & 23.20 & 23.01 & 22.78 & 22.55 & 22.30 & 22.04
					\tabularnewline
					\hline

					\multirow{3}{*}{\tabincell{l}{DnCNN~\cite{ref86_DnCNN}\\+DCLS~\cite{ref82_DCLS}}}
					& \multirow{3}{*}{\tabincell{l}
     {650K\\+19.05M}}  & \multirow{3}{*}{170ms} 
                    & 0 
                    & 24.85 & 24.78 & 24.68 & 24.52 & 24.41 & 24.25 & 24.08 & 23.92 & 23.65 
                    
					\tabularnewline
					&  &  
                        & 10
                        & 23.83 & 23.75 & 23.60 & 23.42 & 23.22 & 23.01 & 22.77 & 22.49 & 22.15
					\tabularnewline
					&  & 
                        & 20 
                        & 23.45 & 23.32 & 23.14 & 22.94 & 22.70 & 22.47 & 22.23 & 22.00 & 21.78
					\tabularnewline
					\hline

\multirow{3}{*}{\tabincell{l}{DASR~\cite{ref3_dasr}}}
					& \multirow{3}{*}{5.84M} & \multirow{3}{*}{49ms} 
                    & 0 
                    & 25.00 & 24.90 & 24.80 & 24.77 & 24.71 & 24.64 & 24.58 & 24.47 & 24.30
                    
					\tabularnewline
					&  &  
                        & 10
                        & 24.07 & 23.93 & 23.77 & 23.56 & 23.37 & 23.20 & 23.02 & 22.82 & 22.63
					\tabularnewline
					&  & 
                        & 20 
                        & 23.33 & 23.18 & 23.02 & 22.84 & 22.66 & 22.48 & 22.30 & 22.12 & 21.95
					\tabularnewline
					\hline

					 \multirow{3}{*}{\tabincell{l}
     {MRDA~\cite{ref6_mrda}}}
					& \multirow{3}{*}{5.84M} & \multirow{3}{*}{57ms} 
                        & 0 
                    & 25.43 & 25.38 & 25.29 & 25.19 & 25.13 & 25.03 & 24.93 & 24.74 & 24.52
					\tabularnewline
					&  & & 10
                        & 24.39 & 24.27 & 24.11 & 23.90 & 23.70 & 23.52 & 23.32 & 23.11 & 22.89
					\tabularnewline
					&  & & 20 
                        & 23.57 & 23.44 & 23.28 & 23.10 & 22.93 & 22.75 & 22.56 & 22.38 & 22.20
					\tabularnewline
					\hline

     \multirow{3}{*}{\tabincell{l}
     {KDSR~\cite{ref5_kdsr}}}
					& \multirow{3}{*}{5.80M} & \multirow{3}{*}{63ms} 
                        & 0 
                    & 25.69 & 25.68 & 25.63 & 25.58 & 25.54 & 25.47 & 25.37 & 25.25 & 25.09
					\tabularnewline
					&  & & 10
                        & 24.58 & 24.48 & 24.33 & 24.13 & 23.93 & 23.75 & 23.54 & 23.32 & 23.12
					\tabularnewline
					&  & & 20 
                        & 23.69 & 23.57 & 23.42 & 23.24 & 23.06 & 22.87 & 22.68 & 22.49 & 22.31
					\tabularnewline
					\hline

                \multirow{3}{*}{ReDSR (Ours)}
					& \multirow{3}{*}{5.79M} & \multirow{3}{*}{43ms} & 0 
                        & \textbf{26.12} & \textbf{26.11} & \textbf{26.07} & \textbf{26.04} & \textbf{26.00} & \textbf{25.90} & \textbf{25.78} & \textbf{25.64} & \textbf{25.44}
					\tabularnewline
					&  & & 10 
                        & \textbf{25.02} & \textbf{24.91} & \textbf{24.78} & \textbf{24.58} & \textbf{24.37} & \textbf{24.17} & \textbf{23.95} & \textbf{23.70} & \textbf{23.47}
					\tabularnewline
					&  & & 20 
                        & \textbf{24.04} & \textbf{23.91} & \textbf{23.77} & \textbf{23.57} & \textbf{23.38} & \textbf{23.17} & \textbf{22.95} & \textbf{22.73} & \textbf{22.52}
					\tabularnewline
					\hline
			\end{tabular}}
		\end{center}
    \label{compareonaniso}
    \vspace{-10pt}
\end{table*}

\begin{figure*}[!t]
		\centering
		\includegraphics[width=1\linewidth]{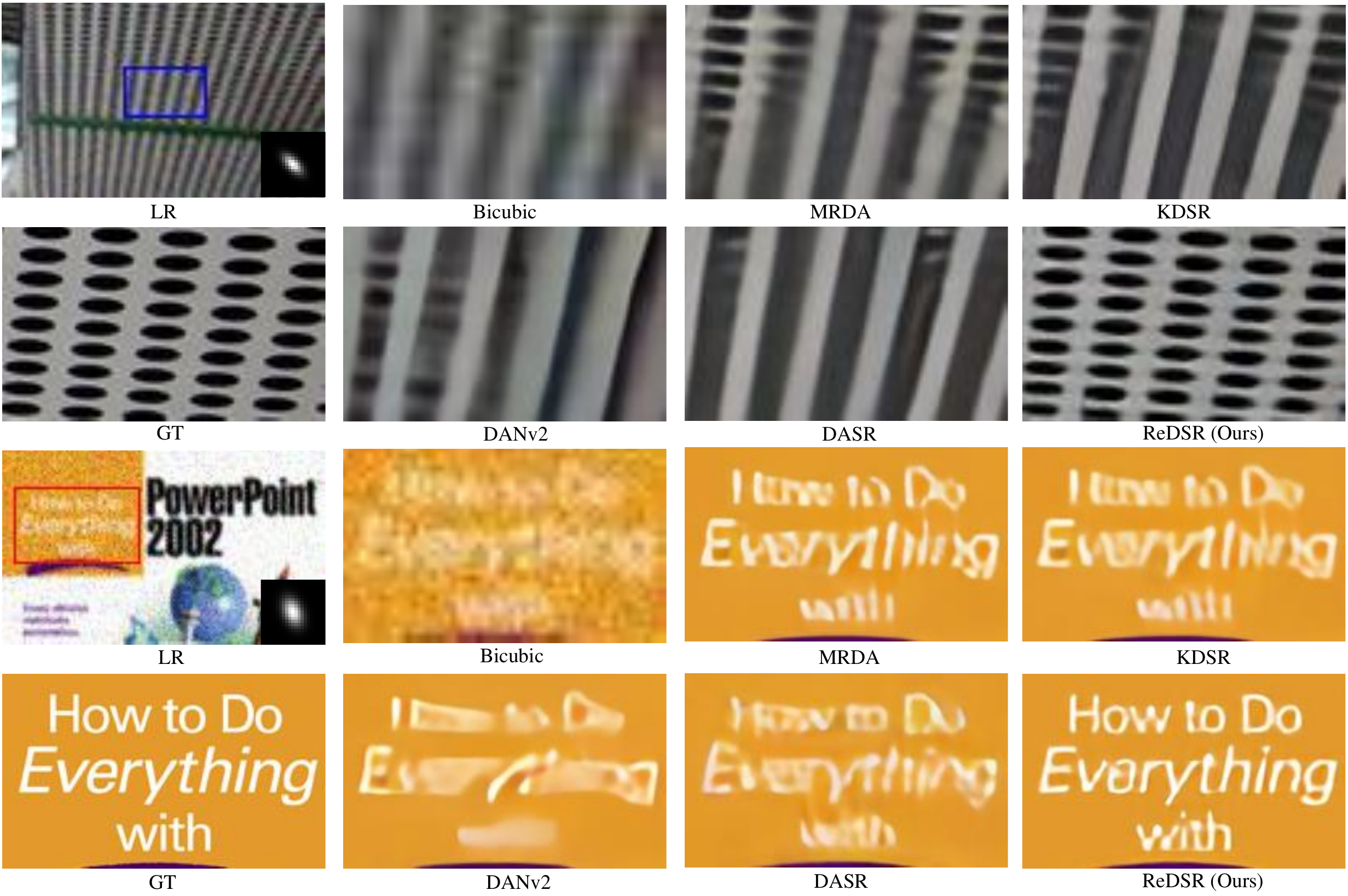}
		\caption{Visual comparison achieved on Urban100 and Set14. Noise level is set to 10 and 20 for these two images, respectively.}
		\label{visual_comparison_aniso}
   \vspace{-15.5pt}
\end{figure*}

\textbf{Visualization of Degradation Representation Distribution:} Our ReDSR employs a distribution alignment loss to associate the distribution of the learned representations with a pre-defined distribution. This bounded constraint enables the learning of a compact representation space with a bijective mapping to the degradation space. To demonstrate this, we synthesize LR images with 6 different degradations (random blur kernels and Gaussian noise levels) and visualize their representations in Fig.~\ref{scatter3}. In addition, we also randomly sample 1000 representations from our pre-defined distribution (\emph{i.e.}, Gaussian distribution). Moreover, we include 3 typical degradation representation learning methods (\emph{i.e.}, DASR~\cite{ref3_dasr}, MRDA~\cite{ref6_mrda} and KDSR~\cite{ref5_kdsr}) for comparison. Without bounded constraint, the representations extracted by previous methods collapse into a small subspace of the Gaussian distribution. In contrast, the representations learned by our ReDSR span the pre-defined distribution, establishing a bijective mapping and facilitating the learning of distinctive representations.

\textbf{(2) Degrader}

\textbf{Re-degradation Loss $\mathcal{L}_{RD}$:} The main idea of this paper is to extract degradation information by re-producing the LR images. To achieve this, the core design is the re-degradation branch and the loss $\mathcal{L}_{RD}$. To validate their effectiveness, we first introduce a model variant (D1) by removing re-degradation branch (\emph{i.e.}, degrader). It can be observed from Table~\ref{tab_ablation1} that the re-degradation branch significantly improves the performance of D1 on all degradations. With our re-degradation branch, our baseline can extract accurate degradation information by reproducing the LR images such that higher PSNR scores are produced. We also visualize the representations extracted from D1 in Fig.~\ref{scatter1}. As we can see, D1 cannot distinguish different degradations in the representation space.

\textbf{Degradation-condition Contamination Injection:} The DAMs and noise injection modules enable the degrader to synthesize pseudo LR images conditioned on degradation representation. To demonstrate their effectiveness, we introduce a network variant (D2), which stretches and concatenates degradation representations with image features~\cite{ref15} before being fed to vanilla convolutions. Then, we develop the network variant D3 by adding noise injection modules to test their effectiveness. It can be observed from Table~\ref{tab_ablation1} that D3 outperforms D2 with significant improvements in PSNR. To further improve the diversity of synthetic pseudo LR images, we replace vanilla convolution on concatenated features by DAM to obtain our baseline. As we can see, our baseline surpasses D3 with notable margins. Without degradation-condition contamination injection, the diversity of synthetic pseudo LR images for D2 is quite limited. As a result, the generator of D2 cannot handle complex degradations during training. With the contamination injection, the degrader of the baseline can reproduce high-quality pseudo LR images conditioned on degradation representations, such that higher PSNR scores are produced.

\subsubsection{Degradation-aware Network Architecture}
\ 
\newline
\indent \textbf{(1) Generator}

\textbf{Degradation-aware Vision State Space Module (DAVSSM):} Within our generator, DAVSSMs are used to achieve a larger efficient receptive field and model long-range dependency. Specifically, to maintain spatial continuity during the scan process, we implement a four-direction zigzag scan method. Besides, a novel DS6 block is proposed to introduce degradation information to state space updating by predicting weighting parameters in SSM from degradation representations. To demonstrate the effectiveness of these two key components, we first introduce a variant (G1) by replacing DAVSSMs with common VSSMs~\cite{ref7_liuvmamba}. Then, we develop another variant G2 by replacing the DS6 block with S6 block~\cite{ref11_MAMBA}.

As shown in Table~\ref{tab_ablation1}, without DAVSSMs, long-range dependency formed by G1 cannot adapt to diverse degradations, such that relatively low performance is produced. With zigzan scan method, G2 can maintain spatial continuity with better performance being achieved. On top of G2, DS6 block can introduce significant performance gain and facilitate our generator to achieve the best results for various degradations. This clearly demonstrates the effectiveness of DAVSSMs.

\textbf{Degradation-condition Modulation Module (DCMM):} To enhance local features and facilitate the expressive power of different channels, we introduce two additional modules (\emph{i.e.}, DAConv and DC$^2$A). To validate their effectiveness, we introduce a variant (G3) by replacing them with vanilla ones. That is, degradation information is not adopted in the DCMM. Then, we develop another variant (G4) by adding DAConv to G3.

As shown in Table~\ref{tab_ablation1}, with DAConv, G4 can effectively enhance the local interaction. In contrast, without using DAConv (\emph{i.e.}, directly employing simple group convolution layers for restoration), G3 can only obtain unsatisfied results, which also supports our previous analysis. Moreover, when introducing degradation information to channel attention (\emph{i.e.}, adding DC$^2$A to G4), our baseline introduce further performance gain to achieve the satisfied results for different degradations.

\begin{figure}[t]
		\centering
		\includegraphics[width=1.0\linewidth]{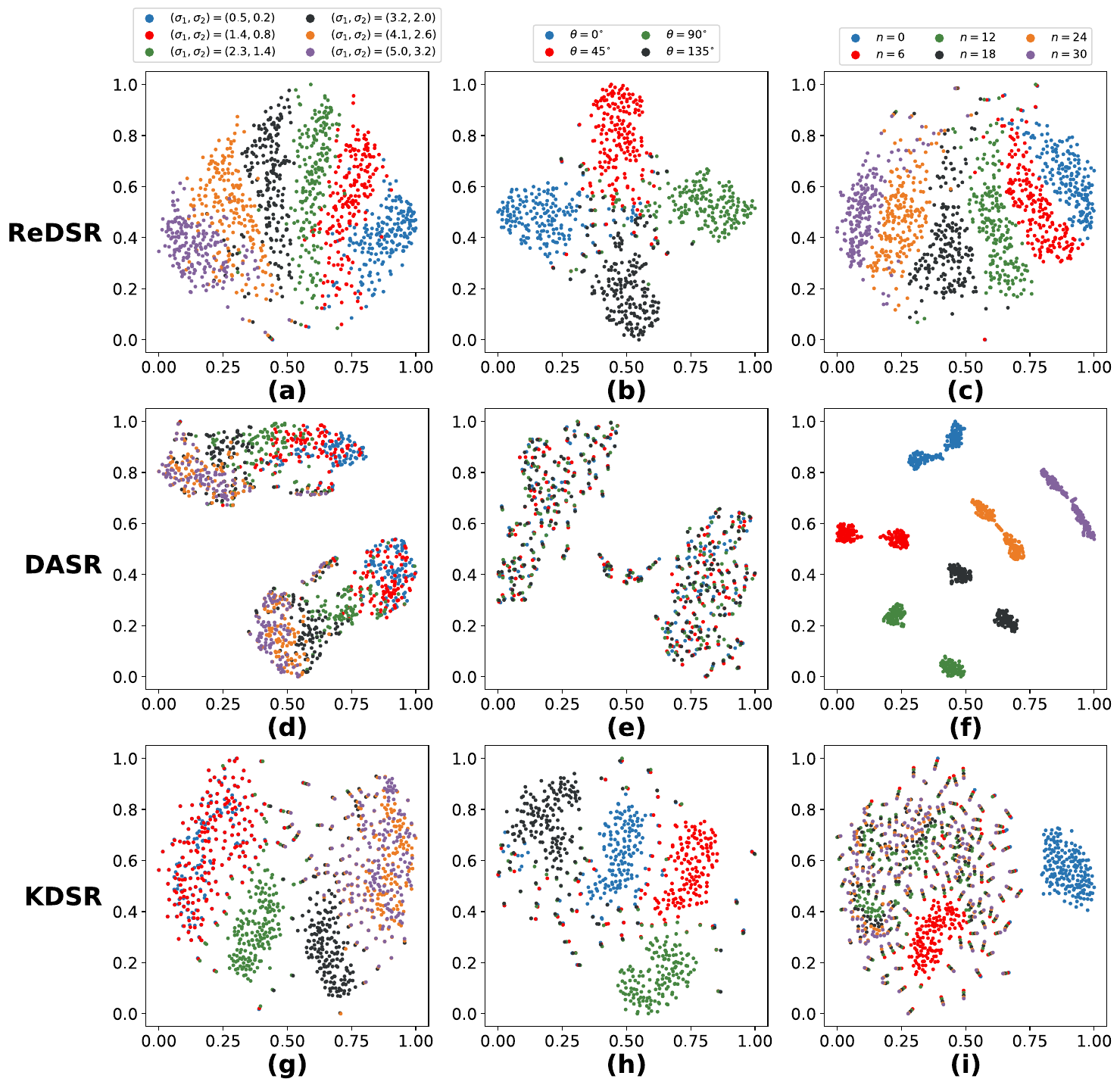}
		\caption{Visualization of representations extracted from images with various blur kernels and noise levels. (a)(d)(g): $\theta=45^{\circ},n=10$; (b)(e)(h): $(\sigma_1,\sigma_2)=(2.4,1.0),n=10$; (c)(f)(i): $(\sigma_1,\sigma_2)=(2.4,1.0),\theta=45^{\circ}$.}
		\label{scatter4}
\end{figure}

\subsection{Performance Evaluation}

\subsubsection{Experiments on Simple Degradation} We first compare the models trained with only isotropic Gaussian kernels. We compare our ReDSR to recent state-of-the-art blind SR methods, including MANet\footnote{Since MANet~\cite{ref38_MANet} is trained on anisotropic Gaussian kernels, we re-train it on isotropic Gaussian kernels for fair comparison.}~\cite{ref38_MANet}, CDSR\footnote{Since the pre-trained model of CDSR~\cite{ref51_CDSR} is unavailable, we re-train it using the officially released codes.}~\cite{ref51_CDSR}, DARSR~\cite{ref83_DARSR}, DAN~\cite{ref37_DAN}, DCLS~\cite{ref82_DCLS}, DASR~\cite{ref3_dasr}, CMDSR~\cite{ref80_CMDSR}, MRDA~\cite{ref6_mrda} and KDSR~\cite{ref5_kdsr}. MANet and DAN require the degradation information as supervision to estimate the degradation model of the LR image. Other methods extract degradation information from the LR images in a fully unsupervised manner. Quantitative results are listed in Table \ref{comparisonwsotas_iso}, with visualization results being provided in Fig.~\ref{visual_comparison_iso}.

\textbf{Quantitative Results:} From Table \ref{comparisonwsotas_iso} we can see that our ReDSR achieves the best performance. Degradation estimation based methods (MANet and DAN) require numerous iterations to achieve accurate estimation of the degradation and suffer relatively long inference time. In contrast, other methods achieve higher efficiency as they employ implicit degradation representation. As compared to DASR, our ReDSR produces significant accuracy improvements. This is because our re-degradation mechanism can better preserve degradation details such that superior performance is achieved. 

\textbf{Qualitative Results:} Figure~\ref{visual_comparison_iso} further compares the visualization results produced by different methods. As we can see, our ReDSR produces results with the best perceptual quality and clearer textures, whereas other methods commonly suffer blurring artifacts and lack of details. This further demonstrates that our Mamba-based backbone is an efficient approach to build long-range dependency while maintaining inference efficiency.

\textbf{Visualization of Degradation Representation:} To further demonstrate the effectiveness of our degradation representation learning scheme, we visualize the degradation representations extracted by our ReDSR, and compare with the contrastive learning scheme (\emph{i.e.}, DASR~\cite{ref3_dasr}). It can be observed from Fig.~\ref{scatter4_iso} that our degradation representation can better distinguish subtle degradation differences, even for unseen degradations in the training set (\emph{e.g.}, representations for $\sigma=4.1$ and $\sigma=5.0$).

\begin{table*}[t]
		\caption{Quantitative comparison on real-world SR competition benchmarks. Inference time is calculated using 256$\times$256 LR images on a NVIDIA RTX 3090 GPU.}
  \vspace{-10pt}
        \renewcommand{\arraystretch}{1.1} 
		\label{tab1}
		\begin{center}
			\small
            \resizebox{1\hsize}{!}{
                \scriptsize
				\begin{tabular}{|clcc|ccc|ccc|}
					\hline 
					 & & &
     
					& \multicolumn{3}{c|}{\multirow{1}{*}{AIM2019-Track2~\cite{ref92_aim2019}}} 
					& \multicolumn{3}{c|}{\multirow{1}{*}{NTIRE2020-Track1~\cite{ref93_ntire2020}}} 
					\tabularnewline
					
					& \multicolumn{1}{l}{Methods} & \#Params. & Time (ms)
					& PSNR $\uparrow$ & SSIM $\uparrow$ & LPIPS $\downarrow$
                    & PSNR $\uparrow$ & SSIM $\uparrow$ & LPIPS $\downarrow$
					\tabularnewline
					\hline
					
					\multirow{6}{*}{\rotatebox{90}{Unpaired}} 
                    & PCR-ESRGAN~\cite{ref90_pcresrgan}
                    & 16.7M & 231
					& 21.59 & 0.610 & 0.321
                    & 24.97 & 0.682 & 0.223
					\tabularnewline

                    & DictSR~\cite{ref89_dictsr} & 8.3M & 89
					& \textcolor{blue}{22.46} & \textcolor[RGB]{202,12,22}{0.629} & \textcolor[RGB]{202,12,22}{0.259}
                    &\textcolor[RGB]{202,12,22}{25.90} & \textcolor[RGB]{202,12,22}{0.711} & \textcolor[RGB]{202,12,22}{0.204}
					\tabularnewline

                    & DAP~\cite{ref95_dap}
                    & 16.9M & 233
					& \textcolor[RGB]{202,12,22}{22.60} & \textcolor{blue}{0.622} & 0.340
                    & 25.40 & 0.707 &  0.252
					\tabularnewline

                    & UDASR~\cite{ref94_udasr}
                    & 16.7M & 231
					&  21.60 & 0.564 & 0.336 
                    & - & - & -
					\tabularnewline

                    & DeFlow~\cite{ref96_deflow}
                    & 16.7M & 231
					& 22.28 & 0.621 & \textcolor{blue}{0.280}
                    & \textcolor{blue}{25.87} & \textcolor{blue}{0.710} & \textcolor{blue}{0.218}
					\tabularnewline

                    & FSSR~\cite{ref97_fssr}
                    & 16.7M & 231
					& 20.82 & 0.510 & 0.390
                    & 23.04 & 0.590 & 0.332
					\tabularnewline


     \hline
     
					\multirow{8}{*}{\rotatebox{90}{Paired}} 
                    & Real-ESRGAN~\cite{ref1_realesrgan}
                    & 16.7M & 231
                    & 22.08 & 0.622 & 0.238
                    & 24.68 & 0.687 & 0.251 
					\tabularnewline

                    & BSRGAN~\cite{ref2_bsrgan}
                    & 16.7M & 231
					& 22.47 & 0.623 & 0.299
                    & 24.56 & 0.669 & 0.265
					\tabularnewline

                    & KDSR~\cite{ref5_kdsr}
                    & 18.9M & 166
					& \textcolor{blue}{22.49} & \textcolor{blue}{0.637} & \textcolor{blue}{0.220}
                    & 25.53 & 0.713 & \textcolor{blue}{0.181}
					\tabularnewline

                    & ZSSR~\cite{ref91_zssr}
                    & 0.9M & -
					& 22.21 & 0.603 & 0.639
                    & 24.93 & 0.642 & 0.620
					\tabularnewline

                    & EDANSR~\cite{ref88_edansr}
                    & 8.1M & 85
					& 22.07 & 0.613 & 0.252
                    & \textcolor{blue}{26.18} & \textcolor{blue}{0.715} & 0.207
					\tabularnewline

                    & SeD~\cite{ref98_SeD}
                    & 16.7M & 231
					& 21.87 & 0.568 & 0.417
                    & 22.30 & 0.399 & 0.484
					\tabularnewline

                 & DANv2~\cite{ref81_DANv2}
                    & 4.71M & 152
					& 22.41 & 0.609 & 0.471
                    & 25.15 & 0.671 & 0.554
					\tabularnewline

                \rowcolor{gray!30}
                & ReDSR (Ours)
                & 7.4M & 56
				& \textcolor[RGB]{202,12,22}{23.37} & \textcolor[RGB]{202,12,22}{0.665} &\textcolor[RGB]{202,12,22}{0.202}
                & \textcolor[RGB]{202,12,22}{26.52} & \textcolor[RGB]{202,12,22}{0.743} &\textcolor[RGB]{202,12,22}{0.168}
					\tabularnewline

					\hline	
			\end{tabular}
   }
		\end{center}
  \label{comparisonwsotas_real}
  \vspace{-15pt}
	\end{table*}

\begin{figure*}[!t]
		\centering
		\includegraphics[width=0.95\linewidth]{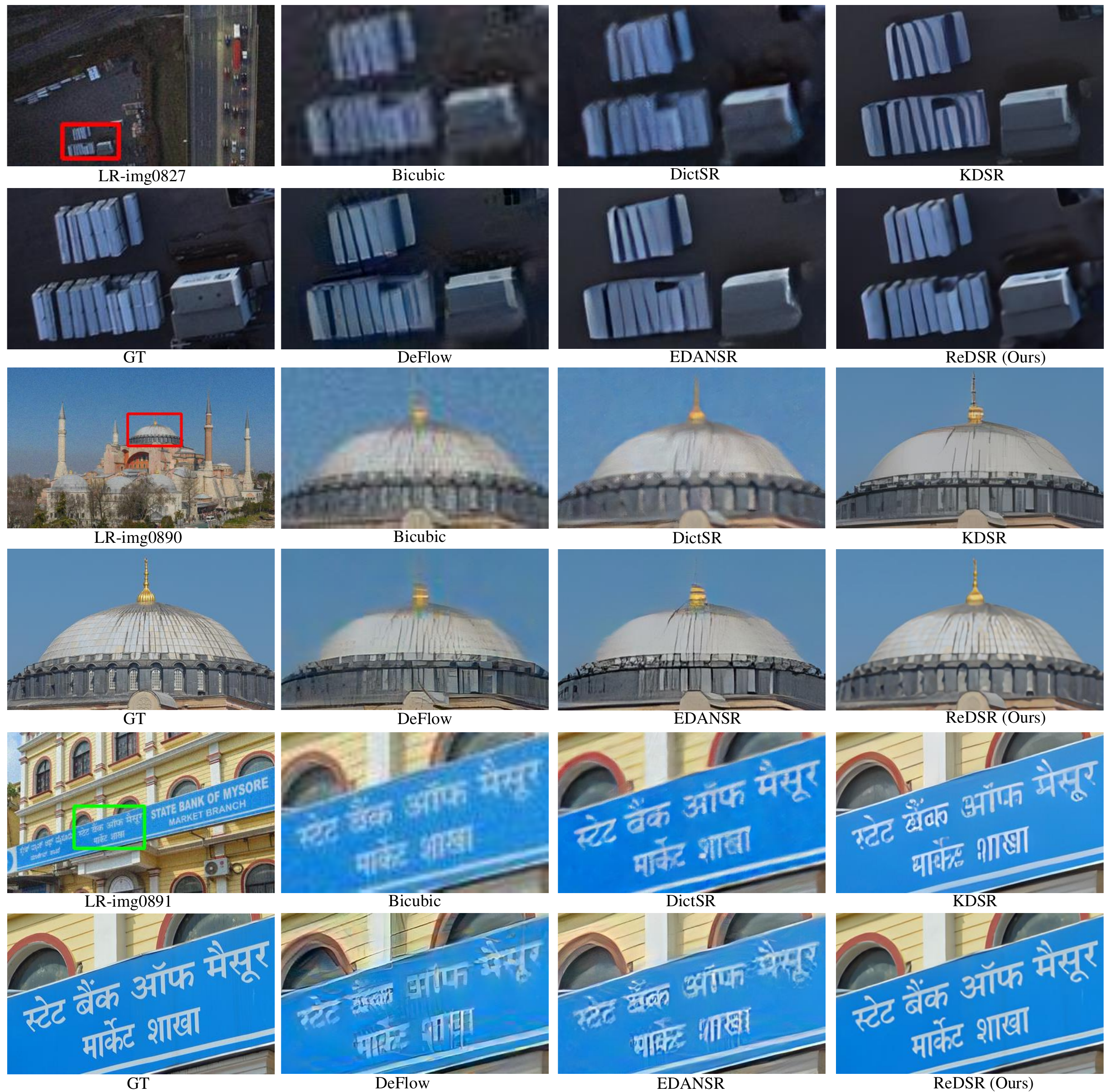}
		\caption{Visual comparison achieved on AIM2019-Track2.}
		\label{visual_comparison_real1}
\end{figure*}

\begin{figure*}[!t]
		\centering
		\includegraphics[width=1.0\linewidth]{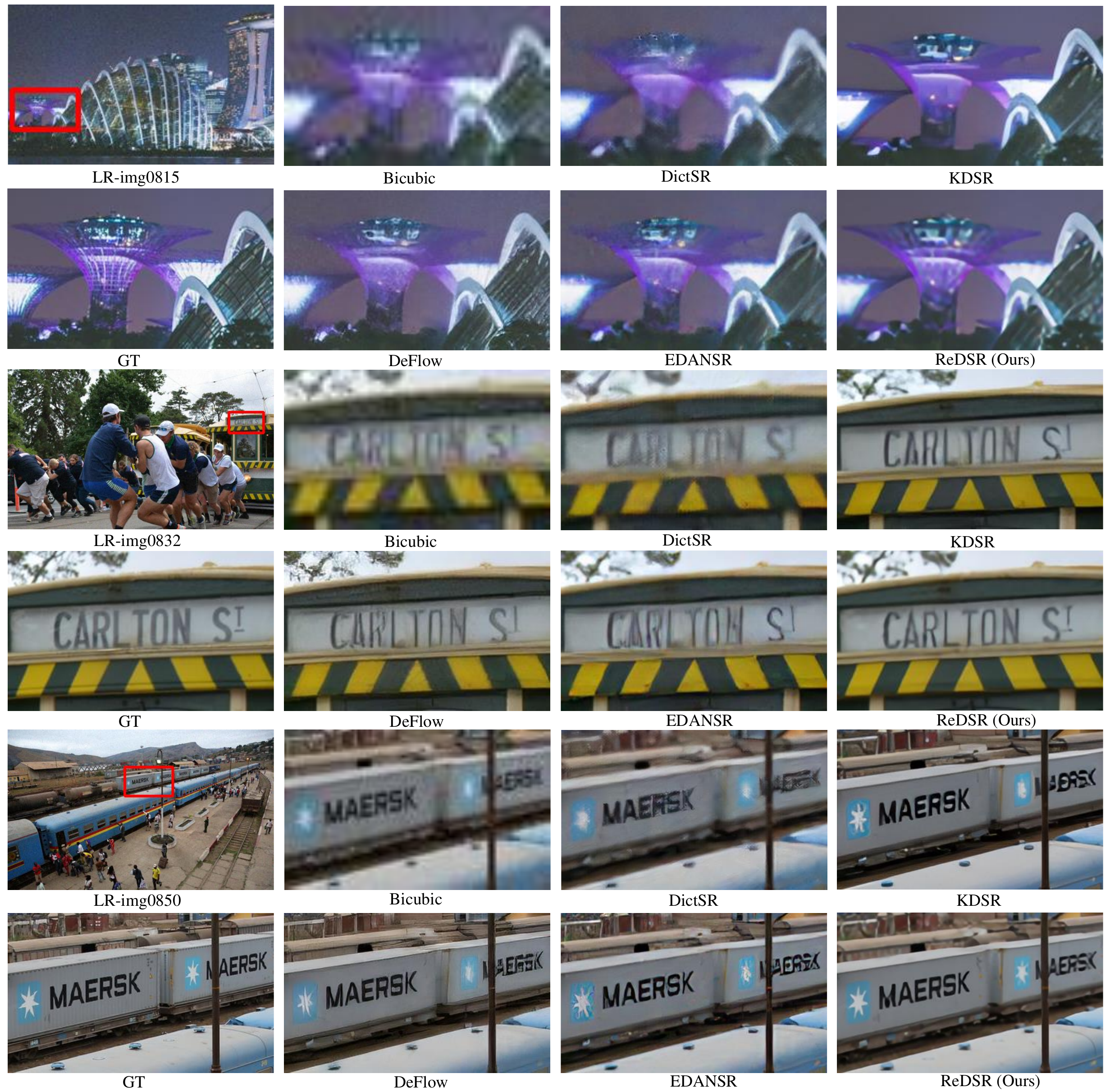}
		\caption{Visual comparison achieved on NTIRE2020-Track1.}
		\label{visual_comparison_real2}
\end{figure*}

\subsubsection{Experiments on General Degradation}

We conduct experiments on general degradations with anisotropic Gaussian kernels and noises. Specifically, 9 anisotropic Gaussian kernels and different noise levels are employed.

\textbf{Quantitative Results.} It can be observed from Table~\ref{compareonaniso} that our ReDSR outperforms other comparative methods on all blur kernels and noise levels. Specifically, DANv2~\cite{ref81_DANv2} performs favorably against another three methods (\emph{i.e.}, DCLS~\cite{ref82_DCLS}, DASR~\cite{ref3_dasr}, MRDA~\cite{ref6_mrda}) but is time-consuming since numerous iterations are required. As compared to DAN, our method produces better performance in terms of both accuracy and efficiency.


\textbf{Qualitative Results:} Visualization results by different methods are illustrated in Fig.~\ref{visual_comparison_aniso}. As we can see, our ReDSR achieves better visual quality with clearer details, while other methods suffer blurring artifacts.


\textbf{Visualization of Degradation Representation:} As compared to simple degradations with only isotropic Gaussian kernels, general degradations are more difficult to be distinguished. We further visualize the degradation representations in Fig.~\ref{scatter4} using the t-SNE method. From the last two rows, we can see that DASR and KDSR cannot well distinguish different degradations. Particularly, degradation representations extracted from images with different noise levels are mixed together in KDSR's representation space, while DASR is confused by different Gaussian kernels. In contrast, our ReDSR produces more distinctive clusters, which demonstrates its effectiveness in extracting accurate degradation information.

\subsubsection{Experiments on Real Degradation}

We further validate the effectiveness of ReDSR on Real-World datasets. As mentioned in Sec.\ref{imple_details}, we train our model with the second-order degradation model in Real-ESRGAN~\cite{ref1_realesrgan}. We conduct experiments on the dataset provided in the challenge of real-world SR (\emph{i.e.}, AIM2019-Track2~\cite{ref92_aim2019} and NTIRE2020-Track1~\cite{ref93_ntire2020}). And PSNR, SSIM, and learned perceptual image patch similarity (LPIPS) are used as metrics to measure real-world SR performance. For performance evaluation, we compare our ReDSR with seven state-of-the-art paired SR methods (Real-ESRGAN~\cite{ref1_realesrgan}, BSRGAN~\cite{ref2_bsrgan}, KDSR~\cite{ref5_kdsr}, ZSSR~\cite{ref91_zssr}, EDANSR~\cite{ref88_edansr}, SeD~\cite{ref98_SeD}, and DANv2~\cite{ref81_DANv2}). And we also include six representative unpaired image SR methods (PCR-ESRGAN~\cite{ref90_pcresrgan}, DictSR~\cite{ref89_dictsr}, DAP~\cite{ref95_dap}, UDASR~\cite{ref94_udasr}, Deflow~\cite{ref96_deflow}, and FSSR~\cite{ref97_fssr}) for reference.

\textbf{Quantitative Results: }Table~\ref{comparisonwsotas_real} presents the quantitative results achieved by different methods. The works focused on SR using unpaired data have difficult to minimize the domain gap between the synthesized LR images and the real-world LR images, which leads to relatively lower quantitative results (\emph{e.g.}, DeFlow and UDASR). In contrast, methods using paired data have leveraged more complicated degradation settings and larger models to produce more promising results on real-world degradations (\emph{e.g.}, BSRGAN and Real-ESRGAN) at a high computational cost and long inference time. Unlike previous paired approaches, our ReDSR extracts discriminative degradation representations and introduces a more advanced vision backbone (\emph{i.e.}, Mamba) to produce better scores among real-world SR methods with fewer parameters.


\textbf{Qualitative Results: }Figure~\ref{visual_comparison_real1} and Figure~\ref{visual_comparison_real2} further compare the visual results produced by different real-world SR methods. It can be observed that previous methods suffer image distortion, obvious artifacts, and lack of details (\emph{e.g.}, the domed roof in the second scene of Fig.~\ref{visual_comparison_real1} and texts in the second scene of Fig.~\ref{visual_comparison_real2}). In contrast, our ReDSR produces visual results with fewer artifacts, finer and more consistent high-frequency details, and higher perceptual quality.

\section{Conclusion}

In this paper, we propose an alternative to learn degradation representations by re-producing LR images. In addition, we introduce a distribution alignment loss to associate learned representations with a pre-defined distribution for superior generalization capability. It is demonstrated that our degradation representation learning scheme can extract discriminative representations to obtain accurate degradation information. Moreover, we introduce a Mamba-based backbone to strike a trade-off between computational efficiency and the global effective receptive field. Specifically, we design a Degradation-aware Vision State Space Module (DAVSSM) with flexible adaption to various degradations based on the degradation representations. Experimental results show that our method achieves state-of-the-art performance for blind SR with various degradations.



\section{Biography Section}
If you have an EPS/PDF photo (graphicx package needed), extra braces are
 needed around the contents of the optional argument to biography to prevent
 the LaTeX parser from getting confused when it sees the complicated
 $\backslash${\tt{includegraphics}} command within an optional argument. (You can create
 your own custom macro containing the $\backslash${\tt{includegraphics}} command to make things
 simpler here.)
 
\vspace{11pt}

\bf{If you include a photo:}\vspace{-33pt}
\begin{IEEEbiography}[{\includegraphics[width=1in,height=1.25in,clip,keepaspectratio]{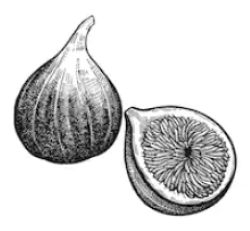}}]{Michael Shell}
Use $\backslash${\tt{begin\{IEEEbiography\}}} and then for the 1st argument use $\backslash${\tt{includegraphics}} to declare and link the author photo.
Use the author name as the 3rd argument followed by the biography text.
\end{IEEEbiography}

\vspace{11pt}

\bf{If you will not include a photo:}\vspace{-33pt}
\begin{IEEEbiographynophoto}{John Doe}
Use $\backslash${\tt{begin\{IEEEbiographynophoto\}}} and the author name as the argument followed by the biography text.
\end{IEEEbiographynophoto}

\vfill


\begin{thebibliography}{1}
\bibliographystyle{IEEEtran}

\bibitem{ref1_realesrgan}
Wang, Xintao, et al. "Real-esrgan: Training real-world blind super-resolution with pure synthetic data." Proceedings of the IEEE/CVF International Conference on Computer Vision. 2021.

\bibitem{ref2_bsrgan}
Zhang, Kai, et al. "Designing a practical degradation model for deep blind image super-resolution." Proceedings of the IEEE/CVF International Conference on Computer Vision. 2021.


\bibitem{ref3_dasr}
Wang, Longguang, et al. "Unsupervised degradation representation learning for blind super-resolution." Proceedings of the IEEE/CVF Conference on Computer Vision and Pattern Recognition. 2021.

\bibitem{ref4}
L. Wang et al., "Unsupervised Degradation Representation Learning for Unpaired Restoration of Images and Point Clouds," in IEEE Transactions on Pattern Analysis and Machine Intelligence 2024.

\bibitem{ref5_kdsr}
Xia, Bin, et al. "Knowledge distillation based degradation estimation for blind super-resolution." ICLR (2023).

\bibitem{ref6_mrda}
Xia, Bin, et al. "Meta-learning-based degradation representation for blind super-resolution." IEEE Transactions on Image Processing 32 (2023): 3383-3396.



\bibitem{ref7_liuvmamba}
Liu, Yue, et al. "VMamba: Visual State Space Model." NIPS (2024).

\bibitem{ref8_zhuvisionmamba}
Zhu, Lianghui, et al. "Vision mamba: Efficient visual representation learning with bidirectional state space model." ICML (2024).

\bibitem{ref9}
Yang, Chenhongyi, et al. "Plainmamba: Improving non-hierarchical mamba in visual recognition." arXiv preprint arXiv:2403.17695 (2024).

\bibitem{ref10}
Zhu, Qinfeng, et al. "Rethinking Scanning Strategies with Vision Mamba in Semantic Segmentation of Remote Sensing Imagery: An Experimental Study." arXiv preprint arXiv:2405.08493 (2024).


\bibitem{ref11_MAMBA}
Gu, Albert, and Tri Dao. "Mamba: Linear-time sequence modeling with selective state spaces." arXiv preprint arXiv:2312.00752 (2023).

\bibitem{ref12}
Hu, Vincent Tao, et al. "Zigma: A dit-style zigzag mamba diffusion model." European Conference on Computer Vision. Springer, Cham, 2025.

\bibitem{ref13_mambair}
Guo, Hang, et al. "Mambair: A simple baseline for image restoration with state-space model." European Conference on Computer Vision. Springer, Cham, 2025.

\bibitem{ref14_ikc}
Gu, Jinjin, et al. "Blind super-resolution with iterative kernel correction." Proceedings of the IEEE/CVF Conference on Computer Vision and Pattern Recognition. 2019.

\bibitem{ref15}
Zhang, Kai, Wangmeng Zuo, and Lei Zhang. "Learning a single convolutional super-resolution network for multiple degradations." Proceedings of the IEEE Conference on Computer Vision and Pattern Recognition. 2018.

\bibitem{ref16_udvd}
Xu, Yu-Syuan, et al. "Unified dynamic convolutional network for super-resolution with variational degradations." Proceedings of the IEEE/CVF Conference on Computer Vision and Pattern Recognition. 2020.


\bibitem{ref17}
He, Jingwen, Chao Dong, and Yu Qiao. "Modulating image restoration with continual levels via adaptive feature modification layers." Proceedings of the IEEE/CVF Conference on Computer Vision and Pattern Recognition. 2019.


\bibitem{ref18_cresmd}
He, Jingwen, Chao Dong, and Yu Qiao. "Interactive multi-dimension modulation with dynamic controllable residual learning for image restoration." European Conference on Computer Vision. Springer International Publishing, 2020.


\bibitem{ref19_perloss}
Johnson, Justin, Alexandre Alahi, and Li Fei-Fei. "Perceptual losses for real-time style transfer and super-resolution." European Conference on Computer Vision. Springer International Publishing, 2016.

\bibitem{ref20_div2k}
Agustsson, Eirikur, and Radu Timofte. "Ntire 2017 challenge on single image super-resolution: Dataset and study." Proceedings of the IEEE Conference on Computer Vision and Pattern Recognition Workshops. 2017.


\bibitem{ref21_flickr2k}
Timofte, Radu, et al. "Ntire 2017 challenge on single image super-resolution: Methods and results." Proceedings of the IEEE Conference on Computer Vision and Pattern Recognition Workshops. 2017.


\bibitem{ref22_manga109}
Matsui, Yusuke, et al. "Sketch-based manga retrieval using manga109 dataset." Multimedia Tools and Applications 76 (2017): 21811-21838.


\bibitem{ref23_set5}
Bevilacqua, Marco, et al. "Low-complexity single-image super-resolution based on nonnegative neighbor embedding." (2012): 135-1.


\bibitem{ref24_set14}
Zeyde, Roman, Michael Elad, and Matan Protter. "On single image scale-up using sparse-representations." Curves and Surfaces: 7th International Conference, Avignon, France, June 24-30, 2010, Revised Selected Papers 7. Springer Berlin Heidelberg, 2012.

\bibitem{ref25_B100}
Martin, David, et al. "A database of human segmented natural images and its application to evaluating segmentation algorithms and measuring ecological statistics." Proceedings eighth IEEE International Conference on Computer Vision. ICCV 2001. Vol. 2. IEEE, 2001.


\bibitem{ref26_Urban100}
Huang, Jia-Bin, Abhishek Singh, and Narendra Ahuja. "Single image super-resolution from transformed self-exemplars." Proceedings of the IEEE Conference on Computer Vision and Pattern Recognition. 2015.

\bibitem{ref27_adam}
Kingma, Diederik P. "Adam: A method for stochastic optimization." arXiv preprint arXiv:1412.6980 (2014).



\bibitem{ref28_outdoortraining}
Wang, Xintao, et al. "Recovering realistic texture in image super-resolution by deep spatial feature transform." Proceedings of the IEEE Conference on Computer Vision and Pattern Recognition. 2018.

\bibitem{ref29_Hellingerdistance1}
González-Castro, Víctor, Rocío Alaiz-Rodríguez, and Enrique Alegre. "Class distribution estimation based on the Hellinger distance." Information Sciences 218 (2013): 146-164.

\bibitem{ref30_Hellingerdistance2}
Wu, Yuefeng, and Giles Hooker. "Hellinger distance and Bayesian non-parametrics: hierarchical models for robust and efficient Bayesian inference." arXiv preprint arXiv:1309.6906 (2013).


\bibitem{ref31_Encyclopedia}
Deza, Elena, et al. Encyclopedia of distances. Springer Berlin Heidelberg, 2009.

\bibitem{ref32_internalgan}
Bell-Kligler, Sefi, Assaf Shocher, and Michal Irani. "Blind super-resolution kernel estimation using an internal-gan." Advances in Neural Information Processing Systems 32 (2019).


\bibitem{ref33_SRCNN}
Dong, Chao, et al. "Image super-resolution using deep convolutional networks." IEEE Transactions on Pattern Analysis and Machine Intelligence 38.2 (2015): 295-307.


\bibitem{ref34_RCAN}
Lim, Bee, et al. "Enhanced deep residual networks for single image super-resolution." Proceedings of the IEEE Conference on Computer Vision and Pattern Recognition Workshops. 2017.

\bibitem{ref35_SRRRN}
Tai, Ying, Jian Yang, and Xiaoming Liu. "Image super-resolution via deep recursive residual network." Proceedings of the IEEE Conference on Computer Vision and Pattern Recognition. 2017.



\bibitem{ref36_VDSR}
Zhang, Yulun, et al. "Image super-resolution using very deep residual channel attention networks." Proceedings of the European Conference on Computer Vision (ECCV). 2018.


\bibitem{ref37_DAN}
Huang, Yan, et al. "Unfolding the alternating optimization for blind super resolution." Advances in Neural Information Processing Systems 33 (2020): 5632-5643.



\bibitem{ref38_MANet}
Liang, Jingyun, et al. "Mutual affine network for spatially variant kernel estimation in blind image super-resolution." Proceedings of the IEEE/CVF International Conference on Computer Vision. 2021.


\bibitem{ref39_contrastive1}
Dosovitskiy, Alexey, et al. "Discriminative unsupervised feature learning with convolutional neural networks." Advances in Neural Information Processing Systems 27 (2014).


\bibitem{ref40_contrastive2}
He, Kaiming, et al. "Momentum contrast for unsupervised visual representation learning." Proceedings of the IEEE/CVF Conference on Computer Vision and Pattern Recognition. 2020.


\bibitem{ref41_KOALANET}
Kim, Soo Ye, Hyeonjun Sim, and Munchurl Kim. "Koalanet: Blind super-resolution using kernel-oriented adaptive local adjustment." Proceedings of the IEEE/CVF Conference on Computer Vision and Pattern Recognition. 2021.

\bibitem{ref42_REBLUR}
Li, Dasong, et al. "Learning degradation representations for image deblurring." European Conference on Computer Vision. Cham: Springer Nature Switzerland, 2022.


\bibitem{ref43_CVAE}
Bao, Jianmin, et al. "CVAE-GAN: fine-grained image generation through asymmetric training." Proceedings of the IEEE International Conference on Computer Vision. 2017.


\bibitem{ref44_SWINIR}
Liang, Jingyun, et al. "Swinir: Image restoration using swin transformer." Proceedings of the IEEE/CVF International Conference on Computer Vision. 2021.


\bibitem{ref45_HAT}
Chen, Xiangyu, et al. "Activating more pixels in image super-resolution transformer." Proceedings of the IEEE/CVF Conference on Computer Vision and Pattern Recognition. 2023.

\bibitem{ref46_hungry}
Fu, Daniel Y., et al. "Hungry hungry hippos: Towards language modeling with state space models." arXiv preprint arXiv:2212.14052 (2022).


\bibitem{ref47_efficientlySSM}
Gu, Albert, Karan Goel, and Christopher Ré. "Efficiently modeling long sequences with structured state spaces." arXiv preprint arXiv:2111.00396 (2021).


\bibitem{ref48_longSSM}
Mehta, Harsh, et al. "Long range language modeling via gated state spaces." arXiv preprint arXiv:2206.13947 (2022).



\bibitem{ref49_simplified}
Smith, Jimmy TH, Andrew Warrington, and Scott W. Linderman. "Simplified state space layers for sequence modeling." arXiv preprint arXiv:2208.04933 (2022).

\bibitem{ref50_umamba}
Ma, Jun, Feifei Li, and Bo Wang. "U-mamba: Enhancing long-range dependency for biomedical image segmentation." arXiv preprint arXiv:2401.04722 (2024).


\bibitem{ref51_CDSR}
Zhou, Yifeng, et al. "Joint learning content and degradation aware feature for blind super-resolution." Proceedings of the 30th ACM International Conference on Multimedia. 2022.



\bibitem{ref52_DAASR}
Wang, Yue, et al. "Blind image super-resolution with degradation-aware adaptation." Proceedings of the Asian Conference on Computer Vision. 2022.

\bibitem{ref53_du2019implicit}
Du, Yilun, and Igor Mordatch. "Implicit generation and modeling with energy based models." Advances in Neural Information Processing Systems 32 (2019).

\bibitem{ref54_hadselldimension}
Hadsell, Raia, Sumit Chopra, and Yann LeCun. "Dimensionality reduction by learning an invariant mapping." 2006 IEEE Computer Society Conference on Computer Vision and Pattern Recognition. Vol. 2. IEEE, 2006.

\bibitem{ref55_kumar}
Kumar, Sanjiv, and Martial Hebert. "Discriminative fields for modeling spatial dependencies in natural images." Advances in Neural Information Processing Systems 16 (2003).

\bibitem{ref56_Lecuntutorial}
LeCun, Yann, et al. "A tutorial on energy-based learning." Predicting Structured Data 1.0 (2006).


\bibitem{ref57_Zhaoenergy}
Zhao, Junbo. "Energy-based Generative Adversarial Network." arXiv preprint arXiv:1609.03126 (2016).



\bibitem{ref58_Hintontraining}
Hinton, Geoffrey, et al. "Unsupervised discovery of nonlinear structure using contrastive backpropagation." Cognitive Science 30.4 (2006): 725-731.


\bibitem{ref59_mnihlearning}
Mnih, Andriy, and Geoffrey Hinton. "Learning nonlinear constraints with contrastive backpropagation." Proceedings. 2005 IEEE International Joint Conference on Neural Networks, 2005.. Vol. 2. IEEE, 2005.

\bibitem{ref60_ranzatoefficient}
Ranzato, Marc'Aurelio, et al. "Efficient learning of sparse representations with an energy-based model." Advances in Neural Information Processing Systems 19 (2006).

\bibitem{ref61_grettonkernel}
Gretton, Arthur, et al. "A kernel method for the two-sample-problem." Advances in Neural Information Processing Systems 19 (2006).




\bibitem{ref62_rizzoenergydistance}
Rizzo, Maria L., and Gábor J. Székely. "Energy distance." Wiley Interdisciplinary Reviews: Computational Statistics 8.1 (2016): 27-38.


\bibitem{ref63_Goldenberg}
Goldenberg, Igor, and Geoffrey I. Webb. "Survey of distance measures for quantifying concept drift and shift in numeric data." Knowledge and Information Systems 60.2 (2019): 591-615.



\bibitem{ref64_generativematching}
Li, Yujia, Kevin Swersky, and Rich Zemel. "Generative moment matching networks." International Conference on Machine Learning. PMLR, 2015.

\bibitem{ref65_Gritsenkospectral}
Gritsenko, Alexey, et al. "A spectral energy distance for parallel speech synthesis." Advances in Neural Information Processing Systems 33 (2020): 13062-13072.



\bibitem{ref66_VAE}
Kingma, Diederik P. "Auto-encoding variational bayes." arXiv preprint arXiv:1312.6114 (2013).



\bibitem{ref67_KLdivergence}
Kullback, Solomon, and Richard A. Leibler. "On information and sufficiency." The Annals of Mathematical Statistics 22.1 (1951): 79-86.


\bibitem{ref68_AckleyBoltzmann}
Ackley, David H., Geoffrey E. Hinton, and Terrence J. Sejnowski. "A learning algorithm for Boltzmann machines." Cognitive Science 9.1 (1985): 147-169.


\bibitem{ref69_Hintoncontrastivedivergence}
Hinton, Geoffrey E. "Training products of experts by minimizing contrastive divergence." Neural Computation 14.8 (2002): 1771-1800.


\bibitem{ref70_SalakhutdinovDeepBoltzmann}
Salakhutdinov, Ruslan, and Geoffrey Hinton. "Deep boltzmann machines." Artificial Intelligence and Statistics. PMLR, 2009.


\bibitem{ref71_SalakhutdinovEfficient}
Salakhutdinov, Ruslan, and Hugo Larochelle. "Efficient learning of deep Boltzmann machines." Proceedings of the Thirteenth International Conference on Artificial Intelligence and Statistics. JMLR Workshop and Conference Proceedings, 2010.


\bibitem{ref72_ZhaoEBGAN}
Zhao, Junbo. "Energy-based Generative Adversarial Network." arXiv preprint arXiv:1609.03126 (2016).


\bibitem{ref73_GAN}
Goodfellow, Ian, et al. "Generative adversarial nets." Advances in Neural Information Processing Systems 27 (2014).


\bibitem{ref74_AE}
Hinton, Geoffrey E., and Ruslan R. Salakhutdinov. "Reducing the dimensionality of data with neural networks." Science 313.5786 (2006): 504-507.



\bibitem{ref75_Vmambair}
Shi, Yuan, et al. "Vmambair: Visual state space model for image restoration." arXiv preprint arXiv:2403.11423 (2024).


\bibitem{ref76_mambaunet2}
Wang, Jinhong, et al. "Large window-based mamba unet for medical image segmentation: Beyond convolution and self-attention." arXiv preprint arXiv:2403.07332 (2024).

\bibitem{ref77_sstransformer}
Islam, Md Mohaiminul, et al. "Efficient movie scene detection using state-space transformers." Proceedings of the IEEE/CVF Conference on Computer Vision and Pattern Recognition. 2023.

\bibitem{ref78_s4nd}
Nguyen, Eric, et al. "S4nd: Modeling images and videos as multidimensional signals with state spaces." Advances in Neural Information Processing Systems 35 (2022): 2846-2861.

\bibitem{ref79_ssvideo}
Wang, Jue, et al. "Selective structured state-spaces for long-form video understanding." Proceedings of the IEEE/CVF Conference on Computer Vision and Pattern Recognition. 2023.


\bibitem{ref80_CMDSR}
Yin, Guanghao, et al. "Conditional hyper-network for blind super-resolution with multiple degradations." IEEE Transactions on Image Processing 31 (2022): 3949-3960.

\bibitem{ref81_DANv2}
Luo, Zhengxiong, et al. "End-to-end alternating optimization for real-world blind super resolution." International Journal of Computer Vision 131.12 (2023): 3152-3169.

\bibitem{ref82_DCLS}
Luo, Ziwei, et al. "Deep constrained least squares for blind image super-resolution." Proceedings of the IEEE/CVF Conference on Computer Vision and Pattern Recognition. 2022.

\bibitem{ref83_DARSR}
Zhou, Hongyang, et al. "Learning correction filter via degradation-adaptive regression for blind single image super-resolution." Proceedings of the IEEE/CVF International Conference on Computer Vision. 2023.





\bibitem{ref84_JSdivergence1}
Menéndez, María Luisa, et al. "The jensen-shannon divergence." Journal of the Franklin Institute 334.2 (1997): 307-318.

\bibitem{ref85_JSdivergence2}
Fuglede, Bent, and Flemming Topsoe. "Jensen-Shannon divergence and Hilbert space embedding." International Symposium on Information Theory, 2004. ISIT 2004. Proceedings.. IEEE, 2004.


\bibitem{ref86_DnCNN}
Zhang, Kai, et al. "Beyond a gaussian denoiser: Residual learning of deep cnn for image denoising." IEEE Transactions on Image Processing 26.7 (2017): 3142-3155.



\bibitem{ref87_tsne}
Van der Maaten, Laurens, and Geoffrey Hinton. "Visualizing data using t-SNE." Journal of Machine Learning Research 9.11 (2008).

\bibitem{ref88_edansr}
Liang, Jie, Hui Zeng, and Lei Zhang. "Efficient and degradation-adaptive network for real-world image super-resolution." European Conference on Computer Vision. Cham: Springer Nature Switzerland, 2022.



\bibitem{ref89_dictsr}
Wang, Longguang, et al. "Learning Coupled Dictionaries from Unpaired Data for Image Super-Resolution." Proceedings of the IEEE/CVF Conference on Computer Vision and Pattern Recognition. 2024.


\bibitem{ref90_pcresrgan}
Romero, Andrés, Luc Van Gool, and Radu Timofte. "Unpaired real-world super-resolution with pseudo controllable restoration." Proceedings of the IEEE/CVF Conference on Computer Vision and Pattern Recognition. 2022.


\bibitem{ref91_zssr}
Shocher, Assaf, Nadav Cohen, and Michal Irani. "“zero-shot” super-resolution using deep internal learning." Proceedings of the IEEE Conference on Computer Vision and Pattern Recognition. 2018.


\bibitem{ref92_aim2019}
Lugmayr, Andreas, et al. "Aim 2019 challenge on real-world image super-resolution: Methods and results." 2019 IEEE/CVF International Conference on Computer Vision Workshop (ICCVW). IEEE, 2019.



\bibitem{ref93_ntire2020}
Lugmayr, Andreas, Martin Danelljan, and Radu Timofte. "Ntire 2020 challenge on real-world image super-resolution: Methods and results." Proceedings of the IEEE/CVF Conference on Computer Vision and Pattern Recognition Workshops. 2020.



\bibitem{ref94_udasr}
Wei, Yunxuan, et al. "Unsupervised real-world image super resolution via domain-distance aware training." Proceedings of the IEEE/CVF Conference on Computer Vision and Pattern Recognition. 2021.



\bibitem{ref95_dap}
Wang, Wei, et al. "Unsupervised real-world super-resolution: A domain adaptation perspective." Proceedings of the IEEE/CVF International Conference on Computer Vision. 2021.



\bibitem{ref96_deflow}
Wolf, Valentin, et al. "Deflow: Learning complex image degradations from unpaired data with conditional flows." Proceedings of the IEEE/CVF Conference on Computer Vision and Pattern Recognition. 2021.

\bibitem{ref97_fssr}
Fritsche, Manuel, Shuhang Gu, and Radu Timofte. "Frequency separation for real-world super-resolution." 2019 IEEE/CVF International Conference on Computer Vision Workshop (ICCVW). IEEE, 2019.


\bibitem{ref98_SeD}
Li, Bingchen, et al. "Sed: Semantic-aware discriminator for image super-resolution." Proceedings of the IEEE/CVF Conference on Computer Vision and Pattern Recognition. 2024.

\bibitem{ref99_VDIR}
Soh, Jae Woong, and Nam Ik Cho. "Variational deep image restoration." IEEE Transactions on Image Processing 31 (2022): 4363-4376.


\end{thebibliography}
\end{document}